\documentclass[11pt]{article}

\usepackage[T1]{fontenc}
\usepackage{newtxtext}
\usepackage[amsthm]{newtxmath}
\usepackage[letterpaper,top=0.72in,bottom=0.78in,left=0.82in,right=0.82in]{geometry}
\usepackage{microtype}
\usepackage{setspace}
\usepackage{authblk}
\usepackage{graphicx}
\usepackage{amsmath}
\usepackage{amsthm}
\usepackage{mathrsfs}
\usepackage{xcolor}
\usepackage{textcomp}
\usepackage{booktabs}
\usepackage{array}
\usepackage{caption}
\usepackage{ragged2e}
\usepackage{titlesec}
\usepackage{placeins}
\usepackage{fancyhdr}
\usepackage{needspace}
\usepackage{xurl}
\usepackage[numbers,sort&compress]{natbib}

\usepackage[colorlinks=true,breaklinks=true]{hyperref}

\makeatletter
\newcommand{\preservedsupplementlabel}[2]{%
  \expandafter\gdef\csname r@#1\endcsname{{#2}{}{}{}{}}%
}
\AtBeginDocument{%
  \preservedsupplementlabel{alg:orthopilot_workflow}{1}%
  \preservedsupplementlabel{alg:oracle_scoring}{2}%
  \preservedsupplementlabel{fig:ext_prospective_data_flow}{E4}%
  \preservedsupplementlabel{fig:ext_tree}{H5}%
  \preservedsupplementlabel{sec:ablation_study}{D}%
  \preservedsupplementlabel{sec:details_datasets}{E}%
  \preservedsupplementlabel{sec:retrospective_dataset_curation}{E.1}%
  \preservedsupplementlabel{sec:baselines}{G}%
  \preservedsupplementlabel{sec:orthobench_design_metrics}{H}%
  \preservedsupplementlabel{sec:reader_study}{I}%
  \preservedsupplementlabel{tab:dataset_availability}{E3}%
  \preservedsupplementlabel{tab:ext1}{H5}%
  \preservedsupplementlabel{tab:ext3}{H7}%
  \preservedsupplementlabel{tab:reader_per_phy}{I10}%
  \preservedsupplementlabel{tab:prospective_cohort_reader_study}{J11}%
  \preservedsupplementlabel{tab:surgical_planning_criteria}{J12}%
  \preservedsupplementlabel{tab:score_interpretation}{J13}%
  \preservedsupplementlabel{tab:nasa_tlx}{J14}%
  \preservedsupplementlabel{tab:physician_documentation}{J15}%
  \preservedsupplementlabel{tab:nursing_workflow}{J16}%
  \preservedsupplementlabel{tab:patient_satisfaction}{J17}%
  \preservedsupplementlabel{tab:communication_quality}{J18}%
  \preservedsupplementlabel{tab:health_information}{J19}%
}
\makeatother

\definecolor{ReportInk}{HTML}{191B1F}
\definecolor{ReportSubtle}{HTML}{42464C}
\definecolor{ReportMuted}{HTML}{666B72}
\definecolor{ReportRule}{HTML}{D7DADE}
\definecolor{ReportFootRule}{HTML}{B8BDC3}
\definecolor{ReportLink}{HTML}{315D73}
\hypersetup{
  linkcolor=ReportInk,
  citecolor=ReportLink,
  urlcolor=ReportLink,
  pdftitle={Evidence-Grounded AI for Musculoskeletal Care},
  pdfauthor={Wenjie Li et al.}
}
\newcommand{\emailicon}{%
  \raisebox{-0.7pt}{\includegraphics[height=6.3pt]{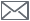}}%
}
\newcommand{\projectfirstpagecontact}{%
  \parbox[b]{\textwidth}{%
    \noindent\color{ReportFootRule}\rule{0.18\textwidth}{0.35pt}\par
    \vspace{3.8pt}%
    \noindent{\sffamily\fontsize{7.4}{9}\selectfont\color{ReportMuted}%
      \textsuperscript{\textdagger}\,\textbf{Project Leader:} Wenjie Li~(Jasper Li)
      \enspace\emailicon\,\,
      \href{mailto:liwenjiesjtu@sjtu.edu.cn}{liwenjiesjtu@sjtu.edu.cn}%
      \enspace\textit{or}\enspace
      \href{mailto:liwenjie@sii.edu.cn}{liwenjie@sii.edu.cn}}%
  }%
}
\fancypagestyle{projectfirstpage}{%
  \fancyhf{}%
  \fancyfoot[C]{%
    \makebox[0pt][c]{%
      \raisebox{14.5pt}[0pt][0pt]{\projectfirstpagecontact}%
    }%
    {\normalfont\normalsize\color{ReportInk}\thepage}%
  }%
}
\titleformat{\section}{\large\bfseries\sffamily\color{ReportInk}}{\thesection}{0.6em}{}
\titleformat{\subsection}{\normalsize\bfseries\sffamily\color{ReportInk}}{\thesubsection}{0.55em}{}
\titleformat{\subsubsection}{\normalsize\bfseries\itshape}{\thesubsubsection}{0.5em}{}
\titlespacing*{\section}{0pt}{2.2ex plus 0.8ex minus 0.3ex}{0.8ex}
\titlespacing*{\subsection}{0pt}{1.8ex plus 0.6ex minus 0.3ex}{0.6ex}
\titlespacing*{\subsubsection}{0pt}{1.5ex plus 0.5ex minus 0.2ex}{0.45ex}

\providecommand{\bibname}{References}
\newcommand{\methodcite}[2]{\mbox{\hyperlink{cite.#1}{[#2]}}}

\theoremstyle{plain}

\theoremstyle{definition}

\theoremstyle{remark}

\newcommand{\paperabstract}{}
\long\def\abstract#1{\gdef\paperabstract{#1}}
\newcommand{\printpaperabstract}{%
  \par{\color{ReportRule}\hrule height 0.45pt}\vspace{7pt}%
  \noindent{\sffamily\bfseries\fontsize{8.3}{9.2}\selectfont\color{ReportInk}\MakeUppercase{Abstract}}\par\vspace{3pt}%
  {\small\justifying\color{ReportInk}\paperabstract\par}\vspace{0.85em}\small%
}

\newcommand{\brandlockup}{%
  \noindent
  \begin{minipage}[c]{0.48\textwidth}
    \includegraphics[height=8.2mm,trim=0 0 470bp 0,clip]{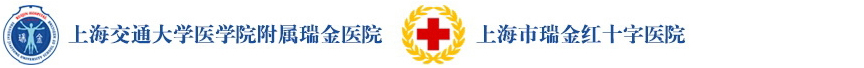}%
  \end{minipage}%
  \hfill
  \begin{minipage}[c]{0.44\textwidth}
    \RaggedLeft
    \includegraphics[height=8.2mm]{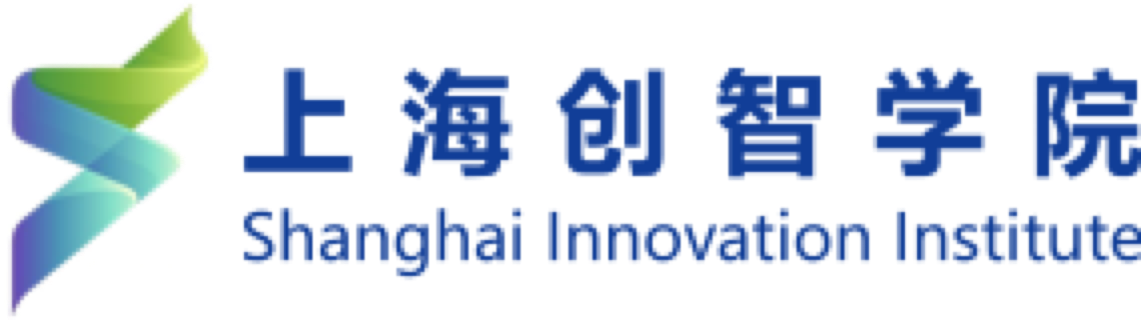}%
  \end{minipage}%
}

\newcommand{\corporateauthors}{%
Wenjie Li\textsuperscript{1,2,3,\textdagger}, Yujie Zhang\textsuperscript{2,3,4}, Fanrui Zhang\textsuperscript{2,5}, Haoran Sun\textsuperscript{3,6}, Renhao Yang\textsuperscript{1}, Junjun He\textsuperscript{2,3}, Weiran Huang\textsuperscript{2}, Yuanfeng Ji\textsuperscript{7}, Chenrun Wang\textsuperscript{2}, Kailing Wang\textsuperscript{8}, Hongcheng Gao\textsuperscript{9}, Kaipeng Zhang\textsuperscript{2}, Hanyu Wang\textsuperscript{1}, Angela Lin Wang\textsuperscript{1}, Xingqi He\textsuperscript{4}, Yilin Huang\textsuperscript{1}, Shiyi Yao\textsuperscript{1}, Lilong Wang\textsuperscript{3}, Yankai Jiang\textsuperscript{3}, Yirong Chen\textsuperscript{3}, Chenglong Ma\textsuperscript{2,3}, Jiyao Liu\textsuperscript{3}, Ming Hu\textsuperscript{3}, Gen Li\textsuperscript{1}, Yidong Xu\textsuperscript{1}, Chengyu Zhuang\textsuperscript{1}, Jiawei Liu\textsuperscript{5}, Yin Zhang\textsuperscript{1}, Lequan Yu\textsuperscript{8}, Lu Chen\textsuperscript{2}, Yinpeng Dong\textsuperscript{9}, Lei Liu\textsuperscript{6}, Carlos Guti{\'e}rrez Sanrom{\'a}n\textsuperscript{10}, Yu Qiao\textsuperscript{2,3}, Weijie Ma\textsuperscript{2}, Xiaosong Wang\textsuperscript{2,3}, Lei Wang\textsuperscript{1}%
}

\newcommand{\frontaffil}[2]{%
  \par\noindent\hangindent=1.55em\hangafter=1%
  \makebox[1.55em][l]{\textsuperscript{#1}}#2\par%
}

\newcommand{\corporateaffiliations}{%
  \frontaffil{1}{Department of Orthopaedics, Ruijin Hospital, College of Health Science and Technology, Shanghai Jiao Tong University School of Medicine, Shanghai, China}%
  \frontaffil{2}{Shanghai Innovation Institute, Shanghai, China}%
  \frontaffil{3}{Shanghai Artificial Intelligence Laboratory, Shanghai, China}%
  \frontaffil{4}{College of Computer Science and Artificial Intelligence, Fudan University, Shanghai, China}%
  \frontaffil{5}{MoE Key Laboratory of Brain-inspired Intelligent Perception and Cognition, University of Science and Technology of China, Hefei, China}%
  \frontaffil{6}{School of Basic Medical Sciences, Intelligent Medicine Institute, Fudan University, Shanghai, China}%
  \frontaffil{7}{Department of Radiation Oncology, Stanford University School of Medicine, Stanford, USA}%
  \frontaffil{8}{School of Computing and Data Science, The University of Hong Kong, Hong Kong SAR, China}%
  \frontaffil{9}{College of AI, Tsinghua University, Beijing, China}%
  \frontaffil{10}{Servicio Cirugia Pediatrica, Department of Pediatric Surgery, Hospital Universitario y Polit{\'e}cnico La Fe, Valencia, Spain}%
}

\makeatletter
\renewcommand{\maketitle}{%
  \begingroup
  \setlength{\parindent}{0pt}%
  \vspace*{-1.55em}%
  \brandlockup\par
  \vspace{6pt}%
  {\color{ReportRule}\hrule height 0.5pt}%
  \vspace{14pt}%
  {\RaggedRight\sffamily\bfseries\fontsize{25}{28}\selectfont\color{ReportInk}%
    Evidence-Grounded AI for\par
    Musculoskeletal Care\par}%
  \vspace{11pt}%
  {\RaggedRight\sffamily\fontsize{8.35}{10.4}\selectfont\color{ReportInk}\corporateauthors\par}%
  \vspace{6pt}%
  {\sffamily\fontsize{6.7}{8.3}\selectfont\color{ReportMuted}\corporateaffiliations}%
  \vspace{5pt}%
  \endgroup
}
\makeatother

\begin{document}

\title{Evidence-Grounded AI for Musculoskeletal Care}

\author[1,2,3]{Wenjie Li}

\author[2,3,4]{Yujie Zhang}

\author[2,5]{Fanrui Zhang}

\author[3,6]{Haoran Sun}

\author[1]{Renhao Yang}

\author[2,3]{Junjun He}

\author[2]{Weiran Huang}

\author[7]{Yuanfeng Ji}

\author[2]{Chenrun Wang}

\author[8]{Kailing Wang}

\author[9]{Hongcheng Gao}

\author[2]{Kaipeng Zhang}

\author[1]{Hanyu Wang}

\author[1]{Angela Lin Wang}

\author[4]{Xingqi He}

\author[1]{Yilin Huang}

\author[1]{Shiyi Yao}

\author[3]{Lilong Wang}

\author[3]{Yankai Jiang}

\author[3]{Yirong Chen}

\author[2,3]{Chenglong Ma}

\author[3]{Jiyao Liu}

\author[3]{Ming Hu}

\author[1]{Gen Li}

\author[1]{Yidong Xu}

\author[1]{Chengyu Zhuang}

\author[5]{Jiawei Liu}

\author[1]{Yin Zhang}

\author[8]{Lequan Yu}

\author[2]{Lu Chen}

\author[9]{Yinpeng Dong}

\author[6]{Lei Liu}

\author[10]{Carlos Guti{\'e}rrez Sanrom{\'a}n}

\author[2,3]{Yu Qiao}

\author[2]{Weijie Ma}

\author[2,3]{Xiaosong Wang}

\author[1]{Lei Wang}

\affil[1]{\enspace Department of Orthopaedics, Ruijin Hospital, College of Health Science and Technology, Shanghai Jiao Tong University School of Medicine, Shanghai, China}

\affil[2]{\enspace Shanghai Innovation Institute, Shanghai, China}

\affil[3]{\enspace Shanghai Artificial Intelligence Laboratory, Shanghai, China}

\affil[4]{\enspace College of Computer Science and Artificial Intelligence, Fudan University, Shanghai, China}

\affil[5]{\enspace MoE Key Laboratory of Brain-inspired Intelligent Perception and Cognition, University of Science and Technology of China, Hefei, China}

\affil[6]{\enspace School of Basic Medical Sciences, Intelligent Medicine Institute, Fudan University, Shanghai, China}

\affil[7]{\enspace Department of Radiation Oncology, Stanford University School of Medicine, Stanford, USA}

\affil[8]{\enspace School of Computing and Data Science, The University of Hong Kong, Hong Kong SAR, China}

\affil[9]{\enspace College of AI, Tsinghua University, Beijing, China}

\affil[10]{\enspace Servicio Cirugia Pediatrica, Department of Pediatric Surgery, Hospital Universitario y Polit{\'e}cnico La Fe, Valencia, Spain}

\date{}

\abstract{
Musculoskeletal diseases are among the leading causes of disability and drive the greatest global need for rehabilitation \cite{WHOMusculoskeletalHealth,cieza2020global}. Because recovery, remodelling and degeneration of bones, joints and related tissues unfold over months to years, care requires longitudinal management rather than isolated decisions \cite{Einhorn2015FractureHealing,Hunter2019Osteoarthritis}. Clinicians must repeatedly integrate evolving patient evidence, medical knowledge and stage-specific functional goals, yet evidence is often fragmented across visits, departments and hospital systems, disrupting continuous, individualised management \cite{Haggerty2003Continuity,WHO2018ContinuityCoordination,Kern2024CareFragmentation}. Here we report OrthoPilot, a clinical artificial intelligence (AI) system powered by a large language model (LLM) that integrates hospital data streams with authoritative external knowledge for continuous musculoskeletal care. It autonomously retrieves real-time imaging, laboratory, pathology and order data and translates evolving patient states into evidence-based decisions from admission diagnosis through rehabilitation planning. We established a specialist-validated benchmark from real-world electronic health records (EHRs) spanning 1,000 disease codes. In a full-pathway reader study against 81 orthopaedic physicians, OrthoPilot outperformed experts with 25 years of experience in diagnostic reasoning, clinical decision-making and management planning. This advantage generalised across 60 external clinical centres, where OrthoPilot surpassed all evaluated intelligent systems. In a prospective physician decision-making study of 1,870 complex cases, OrthoPilot improved full-chain management success by 10.6\%. In a randomised deployment involving 8,240 inpatients, integration into routine care increased cumulative cases per bed by 9.7\% and improved patient-reported access to health information. These results move clinical AI from predicting isolated events toward executing longitudinal management across complete musculoskeletal care pathways.}

\maketitle
\thispagestyle{projectfirstpage}
\printpaperabstract

\section{Introduction}\label{sec1}

\begin{figure}[htbp]
\centering
\includegraphics[width=\textwidth,height=\textheight,keepaspectratio]{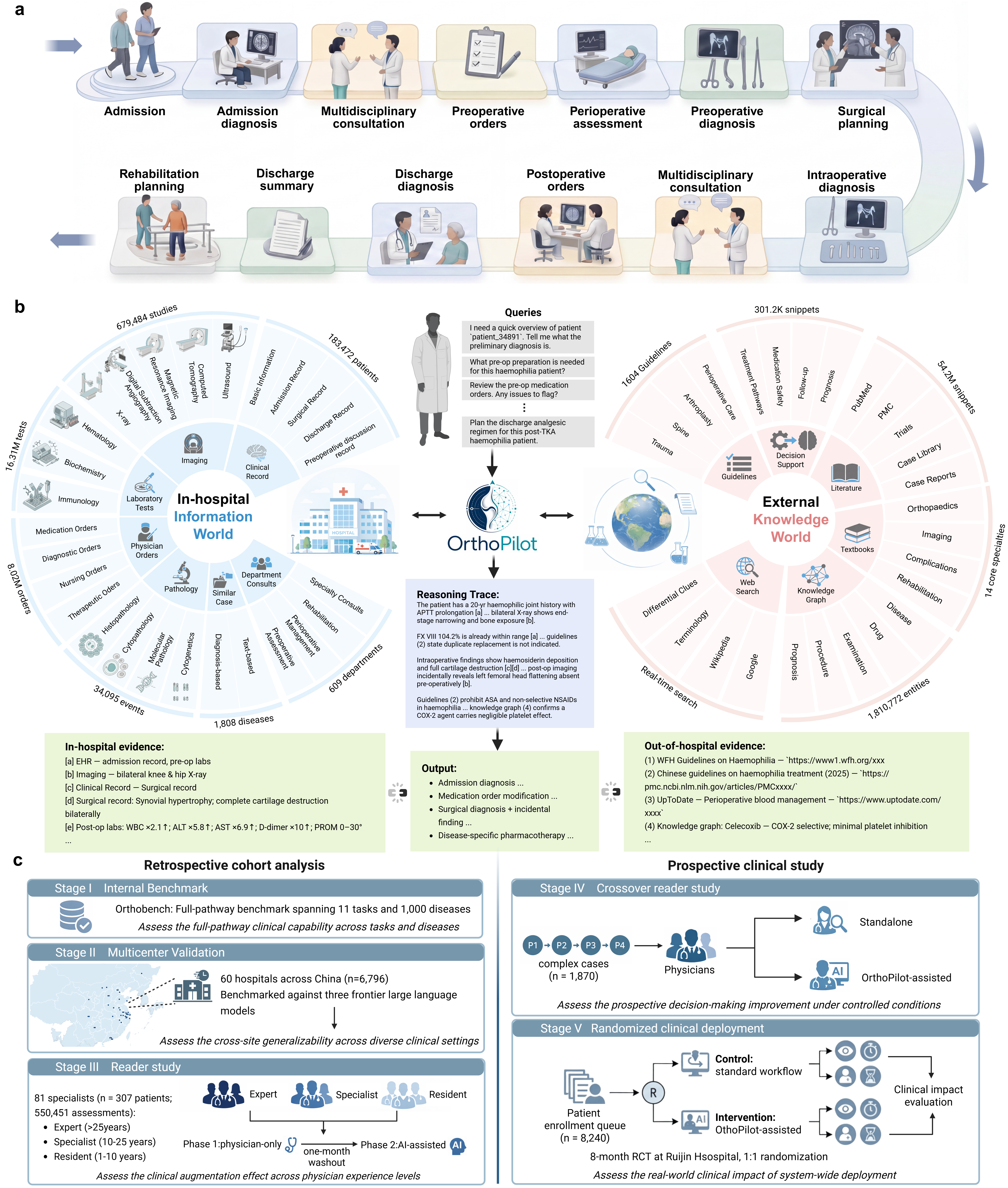}
\caption{\textbf{OrthoPilot: an evidence-grounded agent system for full-pathway musculoskeletal care.} \textbf{a,} Schematic of the musculoskeletal care pathway from admission to rehabilitation, showing sequential diagnostic, perioperative, surgical, discharge, and rehabilitation tasks along time. \textbf{b,} Overview of OrthoPilot and its dual evidence worlds. The system retrieves patient-specific evidence from the in-hospital information world and current medical knowledge from the external knowledge world, then generates a reasoning trace and structured clinical outputs grounded in both evidence sources. \textbf{c,} Retrospective and prospective study design. The retrospective evaluation includes an internal OrthoBench benchmark across 11 tasks and 1,000 diseases, multicentre validation across 60 hospitals and 6,796 cases, and a reader study with 81 physicians, 307 patients, and 550,451 assessments. The prospective evaluation includes a crossover reader study in 1,870 complex cases and an 8-month randomized clinical deployment at Ruijin Hospital with 8,240 enrolled patients.}
\label{fig:introduction}
\end{figure}

Musculoskeletal diseases are a major source of disability and rehabilitation need worldwide. They cause pain, restrict movement and reduce physical function by affecting the body's movement- and load-bearing system \cite{WHOMusculoskeletalHealth,cieza2020global,GBD2023LBP,GBD2023OA}. Unlike many conditions that can be managed around a single diagnosis and a short course of treatment, musculoskeletal diseases are often long-term and dynamic. Pain, structural damage, mobility, treatment response and rehabilitation goals can change over weeks, months or longer \cite{WHOMusculoskeletalHealth,Hunter2019Osteoarthritis,Einhorn2015FractureHealing}. The central challenge in musculoskeletal care is therefore not only to determine what disease a patient has, but to protect or restore function across a prolonged course of treatment and recovery. These features make musculoskeletal care inherently a longitudinal management problem. Here, management denotes the clinical work of converting diagnosis, evolving patient evidence and external medical knowledge into staged decisions for assessment, treatment, discharge and rehabilitation \cite{Cook2018ManagementReasoning,Cook2023ManagementModel,Boyle2025ManagementContext,Lin2020MSKPainGuidelines}. Effective musculoskeletal management must be updated over time and adjusted as the patient's condition changes. Yet this requirement is difficult to meet in routine care. Patient evidence, clinical decisions and care responsibilities are often distributed across encounters, departments and information systems, weakening continuity across the complete care pathway \cite{Bodenheimer2008CareCoordination,Haggerty2003Continuity,Kern2024CareFragmentation,WHO2018ContinuityCoordination}. The core problem is that musculoskeletal management requires long-term, individualized and evidence-based decisions, yet the evidence needed to support these decisions is often fragmented.

Recent progress in clinical AI has created new possibilities for supporting such longitudinal care. Large language models can answer clinical questions, participate in diagnostic dialogue and support differential diagnosis in selected settings \cite{Singhal2023MedPaLM,Tu2025AMIE,McDuff2025DDx}. Agentic systems extend this capacity by linking models to tools and external evidence \cite{Zhao2026DeepRare,Gao2025TxAgent}. Yet these advances still largely address bounded tasks, such as answering a medical question, conducting a diagnostic conversation or selecting a treatment option at one point in care. Recent evaluations also show that LLMs are fragile when they must obtain missing information, follow clinical guidance or operate inside realistic workflows \cite{Hager2024LLMLimitations,Goh2025GPT4PatientCare}. This limitation is critical in musculoskeletal care. Imaging findings, laboratory results, operative observations, medication responses and rehabilitation goals arise at different stages, yet each can alter the next decision. A hospital-deployable system must therefore depend on two tightly coupled capacities. It must first determine what patient evidence remains missing and acquire it from real hospital information systems, including imaging, laboratory results, pathology reports and consultation records. It must then anchor each recommendation to those patient-level facts and to current external medical evidence, rather than to parametric memory alone \cite{Moor2023Foundation,Jiang2023NYUTron}.

Therefore, existing medical AI benchmarks cannot adequately evaluate full-pathway musculoskeletal management. Most benchmarks mainly focus on decontextualised medical question answering, examination-style knowledge tests or isolated diagnostic classification. These evaluations can test whether a model has medical knowledge or can make a diagnostic judgement from a single case fragment. They are less able to show whether a model can maintain consistent, traceable and specialist-aligned reasoning across successive clinical decision points \cite{Singhal2023MedPaLM,Hager2024LLMLimitations}. This gap is especially important in musculoskeletal care, where disease distributions are broad and long-tailed, management tasks can be as clinically consequential as diagnosis, and useful outputs are often open-ended recommendations rather than closed-form answers. Overall, longitudinal musculoskeletal management requires clinical AI to go beyond medical question answering and single-point diagnosis: a useful system must identify missing evidence, acquire patient information, integrate external knowledge and generate traceable recommendations across successive stages of care. Meaningful assessment therefore requires a full-pathway benchmark across diseases and care stages, with specialist-aligned evaluation of open-ended clinical responses.

Here we present OrthoPilot, a hospital-deployable agent system for evidence-grounded longitudinal management in musculoskeletal care (Fig.~\ref{fig:introduction}b). OrthoPilot is not designed to complete isolated medical question answering, but to perform clinically actionable reasoning within real care pathways. Specifically, it identifies missing patient evidence, acquires it from real hospital information systems and verifies recommendations against external medical knowledge. Given a clinical query and patient context, OrthoPilot returns traceable recommendations. OrthoPilot is powered by Clinical Holistic Evidence-to-Execution Synthesis Engine (CHEESE), a lightweight 32-billion-parameter clinical LLM specialised for full-pathway reasoning and trained on expert care trajectories with reinforcement learning signals from sequential decision-making tasks \cite{Guo2025DeepSeekR1,Komorowski2018AIClinician}. The system was built on a 21-year longitudinal cohort from a high-volume tertiary academic hospital in Asia, covering 183,472 patients receiving musculoskeletal care. This data foundation allowed OrthoPilot to learn not only case-level medical judgement, but also how patients move from assessment to treatment, discharge and rehabilitation across real care pathways. At the system-architecture level, OrthoPilot implements a ReAct-style reasoning-and-acting structure \cite{Yao2023ReAct} through the Tool Plaza. In deployment, the system maintains the clinical objective, pathway position, patient state and accumulated trajectory. During reasoning, the system delegates focused evidence acquisition to subagents. These subagents retrieve concise evidence reports from two complementary sources: the in-hospital information world, which provides longitudinal clinical records, imaging, laboratory and pathology results, medical orders, consultation records and analogous cases; and the external knowledge world, which retrieves guidelines, PubMed literature, biomedical knowledge graphs, medical textbooks and real-time web resources. In the example in Fig.~\ref{fig:introduction}b, OrthoPilot integrates hospital records, imaging, surgical findings and haemophilia guidance for a patient with haemophilic arthropathy, then returns evidence-linked diagnostic, medication and treatment recommendations.

To evaluate this capacity for full-pathway clinical management, we first established OrthoBench, which is, to our knowledge, the first clinical-grade benchmark for decision-making across the musculoskeletal care continuum. It was built from 5,905 real patients and spans 1,000 diseases, covering diagnostic reasoning and clinical management. To assess open-ended tasks, we proposed Open Response Assessment for Clinical Language Evaluation (ORACLE), which decomposes each answer into physician-defined decision elements and scores clinically meaningful evidence coverage rather than lexical overlap. We then conducted a comprehensive retrospective and prospective evaluation (Fig.~\ref{fig:introduction}c). Retrospective validation tested OrthoPilot on OrthoBench and evaluated its generalisation on data from 60 external hospitals. A retrospective reader study examined whether OrthoPilot recommendations aligned with specialist judgement and improved physician decision-making. Because prospective evidence for clinical AI remains limited, particularly for systems embedded across complete care pathways \cite{He2019PracticalAI,Rivera2020SPIRITAI,Liu2020CONSORTAI,Vasey2022DECIDEAI,Han2024AIRCT}, we further conducted a crossover reader study and an 8-month randomised deployment to assess sequential decision-making, workflow efficiency, clinical workload and patient-reported experience.

Across retrospective evaluations, OrthoPilot consistently outperformed reasoning LLMs, general-purpose LLMs, medical LLMs and other agent baselines. Its advantage was observed across tasks, in a multicentre dataset from 60 external hospitals, and in both common and rare disease settings. Furthermore, using manually reconstructed patient pathway states, we compared OrthoPilot with 81 physicians across the full pathway. This reader study showed that OrthoPilot reached and exceeded expert-level performance, aligned with specialist judgement and reduced experience-dependent gaps in physician decision-making.

Prospective studies further tested whether these retrospective gains could translate into routine clinical workflows. In these settings, patient evidence was not reconstructed after care, but generated and acquired in real time during clinical practice. In the crossover reader study, OrthoPilot improved decisions at key pathway points and increased full-chain management success from 16.6\% to 27.2\%. In an 8-month randomised deployment, the system improved hospital efficiency, reduced physician and nursing workload, and improved patient-reported communication with clinicians and access to health information. Together, these retrospective and prospective findings suggest that evidence-grounded sequential reasoning can move clinical AI beyond isolated question answering towards longitudinal management in routine musculoskeletal care.

\section{Results}\label{sec2}

\subsection{Overview of the OrthoPilot system}

OrthoPilot is a clinical-ready AI system designed for longitudinal clinical management in musculoskeletal care (Fig.~\ref{fig:introduction}b). At its core is the Clinical Holistic Evidence-to-Execution Synthesis Engine (CHEESE), a 32-billion-parameter LLM trained by reinforcement learning to support sequential decision-making across clinical stages. This setting requires more than an accurate single-step prediction. Management decisions must remain coherent from admission and perioperative planning to discharge and rehabilitation. They must also integrate patient-level evidence from imaging, laboratory tests, pathology, consultations and electronic health records with external knowledge sources that continue to evolve. We therefore designed OrthoPilot around the ``Tool Plaza'', a dual-evidence paradigm that links an in-hospital information world with an external knowledge world. The in-hospital world retrieves and organises patient-specific multimodal evidence, including X-ray, CT and MRI studies. The external world queries knowledge graphs, guidelines, literature, textbooks and web search to ground recommendations in traceable evidence. During application, OrthoPilot realises this paradigm through a hierarchical multi-agent architecture. A CHEESE-driven main agent, equipped with multimodal understanding, defines the reasoning strategy, detects missing evidence in the current clinical context and decomposes complex management questions into focused retrieval subtasks. Specialised sub-agents execute these subtasks across the two evidence worlds and return structured findings. The main agent then summarises these findings into evidence-grounded clinical recommendations. This application-time architecture gives OrthoPilot clinical agency. It allows the system to guide the care pathway, acquire key evidence at each decision point and preserve reasoning consistency across the clinical timeline, rather than passively answering isolated queries.

\subsection{OrthoBench: a full-pathway clinical benchmark for musculoskeletal care}

\FloatBarrier

\begin{figure}[htbp]
\centering
\includegraphics[width=\textwidth,height=\textheight,keepaspectratio]{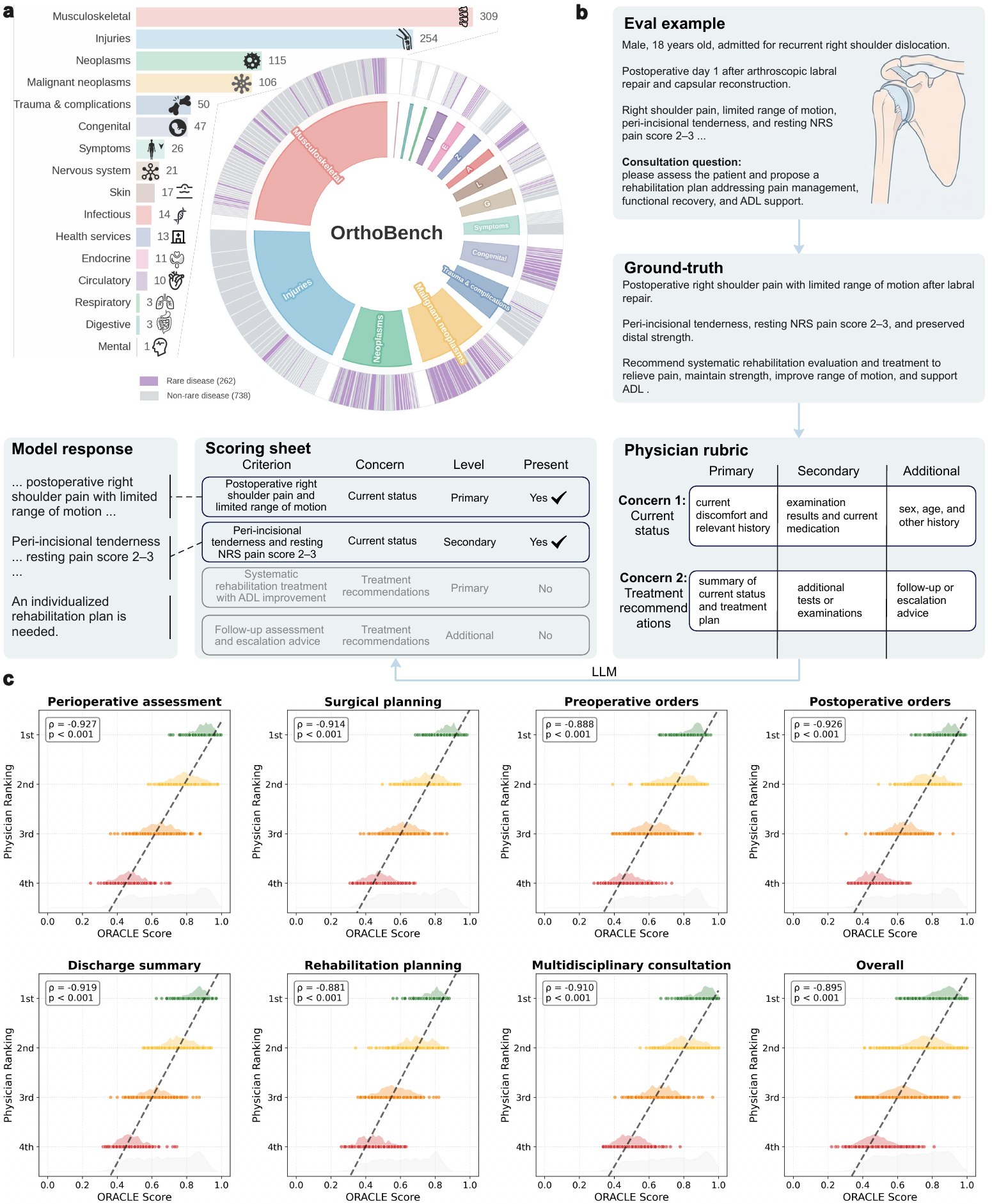}
\caption{\textbf{OrthoBench and ORACLE for full-pathway clinical evaluation.}
\textbf{a,} Overview of OrthoBench disease coverage, showing major disease categories and the distribution of rare and non-rare disease codes.
\textbf{b,} Example ORACLE evaluation workflow for an open-ended clinical management case, showing the clinical query, ground-truth answer, physician rubric, model response and scorecard-based coverage assessment.
\textbf{c,} Validation of ORACLE scores against physician rankings across seven clinical management tasks and overall performance.}
\label{fig:oracle_validation}
\end{figure}

Existing medical AI benchmarks, including broad knowledge tests, medical examination QA, expert-level reasoning tasks and rubric-based health conversations, mostly evaluate LLMs in idealised single-encounter settings \cite{Hendrycks2021MMLU,Jin2021MedQA,Zuo2025MedXpertQA,Arora2025HealthBench}. The model is given a bounded prompt and asked to answer, rather than to acquire missing patient evidence, revise management decisions over time or preserve coherence across a care pathway. To address this gap, we developed OrthoBench as a full-pathway clinical benchmark for musculoskeletal care. It spans the trajectory from presentation to perioperative management, discharge and rehabilitation. OrthoBench was derived from a longitudinal cohort of more than 180,000 patients treated at Ruijin Hospital from 2004 to 2024. It includes 5,905 test patients, 135,745 evaluation instances and 1,000 disease codes across musculoskeletal and trauma conditions (Fig.~\ref{fig:oracle_validation}a, Supplementary Fig.~\ref{fig:ext_tree} and Supplementary Tables~\ref{tab:ext1}--\ref{tab:ext3}). To our knowledge, OrthoBench is the largest clinical benchmark in a surgical speciality designed to test whether LLMs can sustain clinical agency across sequential decision points, rather than pattern-match on decontextualised questions.

The 11 tasks are organised along the clinical timeline. Diagnostic reasoning tasks (Tasks 1--4) comprise Admission diagnosis, Preoperative diagnosis, Intraoperative diagnosis and Discharge diagnosis. These tasks assess diagnostic reasoning as patient evidence accumulates from admission to surgery and discharge. Clinical management tasks (Tasks 5--11) comprise Perioperative assessment, Surgical planning, Preoperative orders, Postoperative orders, Discharge summary, Rehabilitation planning and Multidisciplinary consultation. These tasks evaluate whether a system can generate executable, patient-specific plans across longitudinal care.

To evaluate generalisability, OrthoBench incorporates a two-dimensional stratification scheme. Along the distributional axis, 383 disease codes (3,325 patients) are designated in-distribution (ID), while 617 codes (2,580 patients) are held out as out-of-distribution (OOD). Along the prevalence axis, 262 codes are classified as rare diseases, accounting for 634 patients (10.7\% of the test cohort). The intersection of these axes yields four strata of increasing difficulty: ID non-rare (321 codes, 3,051 patients), ID rare (62 codes, 274 patients), OOD non-rare (417 codes, 2,220 patients) and OOD rare (200 codes, 360 patients). The final stratum, previously unseen rare conditions, provides the most stringent test of clinical reasoning and is intended to remain informative as LLM capabilities improve.

To evaluate open-ended clinical management tasks, we developed the Open Response Assessment for Clinical Language Evaluation framework (ORACLE). These tasks require comprehensive clinical reasoning rather than selection from predefined options. Unlike benchmarks that score free-text outputs by surface overlap, ORACLE decomposes each clinical plan into verifiable decision elements grounded in the patient's medical record (Fig.~\ref{fig:oracle_validation}b). It uses physician-defined rubrics to derive structured scorecards from gold-standard answers and measures whether an LLM response covers clinically salient decision elements across hierarchical requirement levels. We validated ORACLE against physician rankings (Fig.~\ref{fig:oracle_validation}c). Across 691 clinical cases assessed by 12 orthopaedic specialists, ORACLE scores showed strong concordance with physician rankings, with a Spearman's rank correlation coefficient of $-0.895$ across 8,292 model--response evaluations. Task-specific correlations were consistently high across all seven clinical management tasks. The remaining differences were modest and expected, as ORACLE captures explicit clinical content whereas physicians also consider prioritisation, contextual appropriateness and risk sensitivity. To confirm that automated gains reflected clinically meaningful improvements, we further performed blinded specialist ranking and multidimensional assessment of model outputs, including medical accuracy, completeness, clinical safety, actionability and clarity. Detailed benchmark construction and evaluation procedures are provided in Methods Section~\ref{sec:evaluation_framework_benchmark}. These analyses establish ORACLE as a scalable framework for evaluating open-ended clinical management while retaining expert review as a complementary standard.

\subsection{Retrospective cohort analysis}

\subsubsection{Performance across centres}

\FloatBarrier

\begin{figure}[htbp]
\centering
\includegraphics[width=\textwidth,height=\textheight,keepaspectratio]{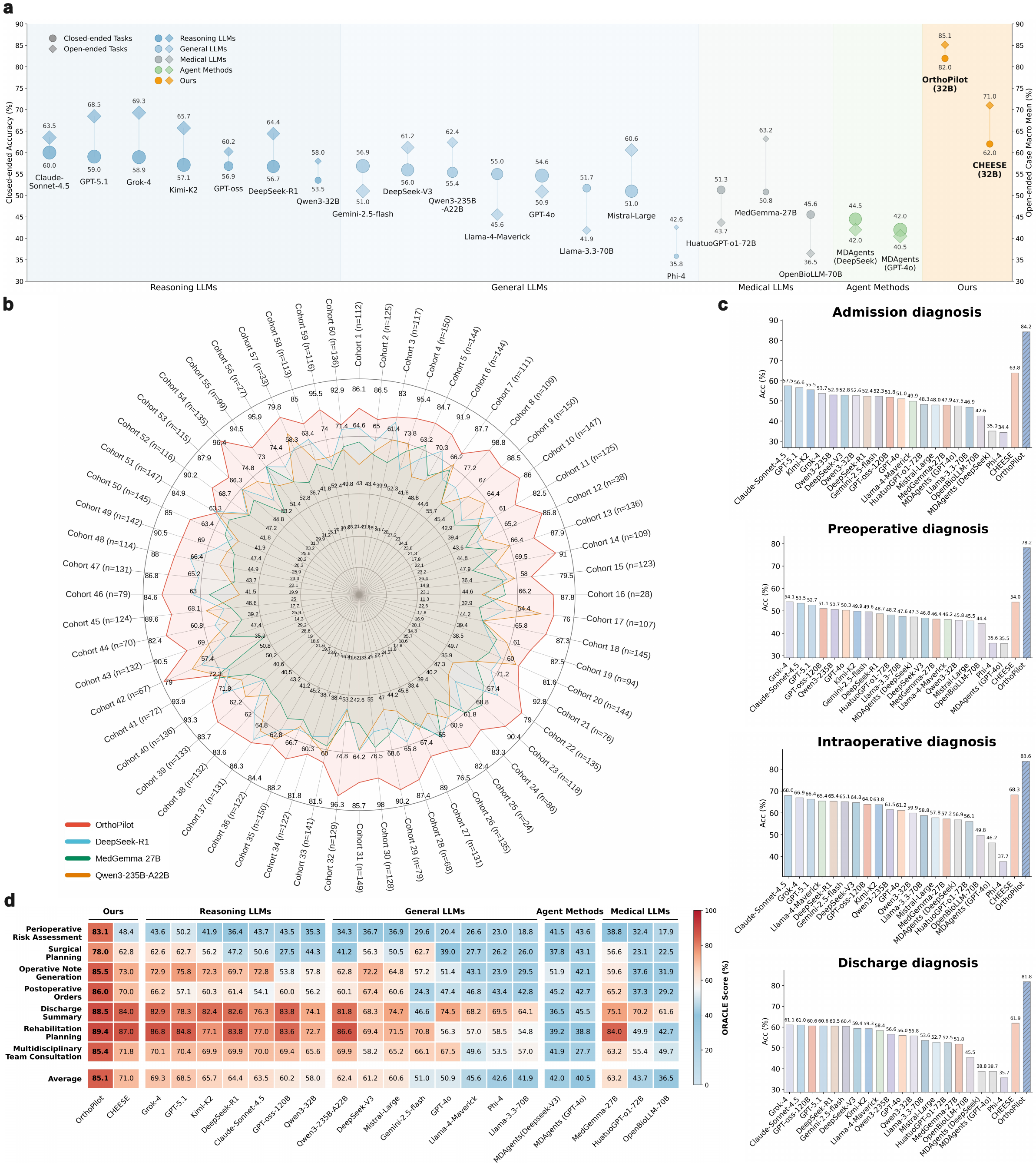}
\caption{\textbf{Aggregate benchmark performance, external generalisability and task-level performance of OrthoPilot.}
\textbf{a,}~Mean performance of 22 systems on OrthoBench, including reasoning LLMs, general-purpose LLMs, medical domain-specific LLMs, agent methods, CHEESE and OrthoPilot. Circles denote closed-ended diagnostic tasks and diamonds denote open-ended clinical management tasks; bubble size is proportional to model parameters, and colours indicate model categories.
\textbf{b,}~Centre-level external validation across 60 hospital centres and 6{,}796 cases, comparing OrthoPilot with three locally deployable open-source baselines: DeepSeek-R1, Qwen3-235B-A22B and MedGemma-27B.
\textbf{c,}~Diagnostic performance at four sequential clinical stages. Bars show task-specific accuracy for each model, with CHEESE and OrthoPilot highlighted.
\textbf{d,}~Open-ended management performance across seven clinical tasks. Colours encode ORACLE scores (\%) from 0 to 100, with warmer colours indicating higher performance; bold values mark the highest score in each row.}
\label{fig:performance_generalisability}
\end{figure}

To test whether AI systems can support longitudinal clinical management rather than isolated prediction, we evaluated performance along two complementary axes of OrthoBench. Closed-ended diagnostic tasks (Tasks~1--4) measured stage-specific accuracy as patient evidence accumulated from admission to discharge. Open-ended clinical management tasks (Tasks~5--11) measured ORACLE scores for executable plans across perioperative assessment, orders, discharge, rehabilitation and consultation. We benchmarked 22 systems across reasoning LLMs, general-purpose LLMs, medical domain-specific LLMs, agent methods, CHEESE and the full OrthoPilot system (Fig.~\ref{fig:performance_generalisability}a; baseline details are provided in Appendix~\ref{sec:baselines}).
The comparison showed that full-pathway clinical management remained difficult for systems built around either general reasoning or domain tuning alone. Reasoning LLMs formed the strongest baseline group, with mean scores of 57.47\% on closed-ended tasks and 64.24\% on open-ended tasks, ahead of general-purpose LLMs, medical domain-specific LLMs and existing agent methods. This pattern suggests that explicit reasoning ability was more useful than biomedical specialization when models had to connect evidence across sequential stages of care. Even the leading baseline models reached only 60.00\% for closed-ended diagnosis and 69.30\% for open-ended management, indicating that frontier reasoning alone did not sustain the clinical coherence required by OrthoBench.
CHEESE shifted this performance ceiling. As a single clinically trained LLM, it ranked second on both evaluation axes, reaching 62.00\% and 71.00\%. This result indicates that reinforcement learning on sequential clinical decision-making produced a stronger reasoning backbone than scale or medical pretraining alone. OrthoPilot further raised performance to 81.95\% and 85.13\%, exceeding CHEESE by 19.95\% and 14.13\%. In contrast, the two MDAgents baselines remained in the low 40\% range. Thus, the gain did not arise from agentic coordination in itself. It emerged when a clinically specialised reasoning backbone was coupled to hierarchical tool use, allowing OrthoPilot to acquire missing evidence and integrate it into management decisions across the care pathway.

External generalizability was examined in an independent multicentre retrospective cohort of 6{,}796 cases from 60 medical centres across different regions of China. Cohort curation details are provided in Appendix~\ref{sec:details_datasets}. Because institutional governance prohibited uploading patient records to third-party APIs, external validation was restricted to locally deployable open-source baselines representing three model families: DeepSeek-R1 (reasoning LLM), Qwen3-235B-A22B (general-purpose LLM) and MedGemma-27B (medical domain-specific LLM).

OrthoPilot retained strong performance across external centres (Fig.~\ref{fig:performance_generalisability}b). Using centre-level unweighted averaging across the 60 hospitals, it achieved a mean closed-ended accuracy of 72.77\% (range 45.97--94.05\%) and a mean open-ended ORACLE score of 79.45\% (range 59.33--96.70\%). It exceeded DeepSeek-R1, Qwen3-235B-A22B and MedGemma-27B by 15.79\%, 18.24\% and 22.10\% on closed-ended tasks, and by 12.04\%, 12.94\% and 13.90\% on open-ended tasks. This advantage was consistent at the institutional level: OrthoPilot outperformed all three baselines in 57--59 of 60 centres for diagnostic tasks and in 50--57 centres for management tasks. Although absolute scores were lower than in the internal benchmark, the same performance hierarchy was preserved across centres. These results indicate that OrthoPilot's gains were not confined to a single hospital distribution, but extended to heterogeneous clinical settings while preserving the distinction between diagnosis and longitudinal management.

\subsubsection{Performance on full-pathway clinical tasks}

Task-level analysis was used to test whether OrthoPilot improved the clinical inflection points at which longitudinal information must be converted into action (Fig.~\ref{fig:performance_generalisability}c,d). Diagnostic tasks provided a chronological test of whether each system could update a patient representation as evidence accrued. For Admission diagnosis, OrthoPilot achieved 84.20\%, the highest score across systems, showing strong early diagnostic performance from incomplete clinical records. The most demanding diagnostic transition occurred before surgery, when admission findings, imaging, laboratory results and surgical indications had to be reconciled before the management plan was fixed. OrthoPilot scored 78.20\% for Preoperative diagnosis, whereas Grok-4 reached 54.10\%. This result identifies the preoperative stage as the point at which longitudinal reasoning most clearly separated OrthoPilot from general reasoning models. Once operative observations became available, diagnostic uncertainty decreased. CHEESE and proprietary LLMs therefore performed strongly on Intraoperative diagnosis, but OrthoPilot remained first at 83.60\%. For Discharge diagnosis, OrthoPilot reached 81.80\%, indicating that it could consolidate the diagnostic sequence into a final record for subsequent management.

The management tasks provided a direct test of full-pathway clinical agency. Perioperative assessment, Surgical planning and Postoperative orders correspond to perioperative optimization, operative strategy and early recovery, which are central components of surgical quality and AI-enabled surgical care \cite{Domenghino2023SurgicalQuality,Varghese2024AISurgery,Odor2020BMJm540}. OrthoPilot scored 83.10\% for Perioperative assessment, 78.00\% for Surgical planning and 86.00\% for Postoperative orders, exceeding the nearest comparator for each task. The same pattern extended beyond the operation. OrthoPilot achieved 88.50\% for Discharge summary, 89.40\% for Rehabilitation planning and 85.40\% for Multidisciplinary consultation, three tasks that depend on care transitions, functional recovery and team-based coordination \cite{Graham2024MappingDischarge,Kashikar2020Comanagement}. In contrast, MDAgents variants remained in the low 40\% range on mean closed-ended and open-ended scores. Thus, the task-level gains were not a simple effect of agentic prompting. They arose when a clinically trained reasoning model was embedded in an architecture able to acquire missing evidence, coordinate specialised tools and preserve management intent across the care pathway.

\subsubsection{Performance on a variety of diseases}

\FloatBarrier

\begin{figure}[htbp]
\centering
\includegraphics[width=\textwidth,height=0.82\textheight,keepaspectratio]{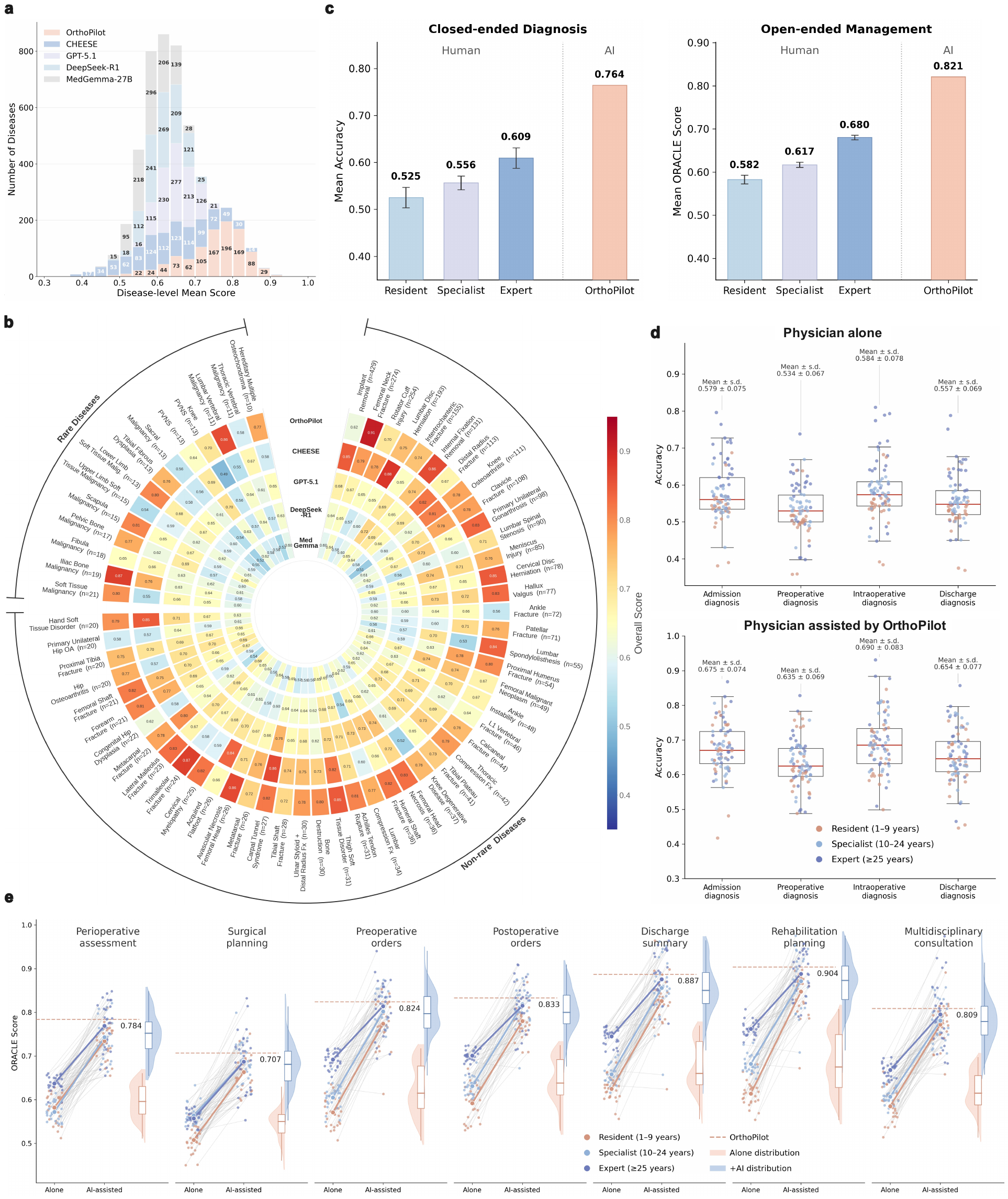}
\caption{\textbf{Disease-level performance and human--AI comparative reader study.}
\textbf{a,}~Stacked histogram of disease-level mean scores for five models across 1{,}000 disease categories. OrthoPilot's distribution is shifted rightward relative to baseline models.
\textbf{b,}~Concentric ring heatmap showing overall scores for 62 diseases with sufficient sample size. Each sector represents one disease ordered by patient count; concentric rings represent models from inner to outer: MedGemma, DeepSeek-R1, GPT-5.1, CHEESE, OrthoPilot.
\textbf{c,}~Mean performance of physicians alone versus OrthoPilot alone in a seven-arm washout-crossover reader study with 81 orthopaedic physicians and 307 patients. Left, mean accuracy on closed-ended diagnostic tasks; right, mean ORACLE score on open-ended management tasks. Physicians are stratified by experience tier. Error bars indicate 95\% confidence intervals.
\textbf{d,}~Per-physician accuracy distributions for four diagnostic tasks under physician-alone and AI-assisted conditions. Each dot represents one physician coloured by experience tier.
\textbf{e,}~Paired slope charts for seven open-ended management tasks. Connecting lines link the same physician across conditions; bold lines indicate tier means. Dashed horizontal lines mark OrthoPilot standalone scores. Half-violin plots compare physician-alone and AI-assisted distributions.}
\label{fig:disease_reader}
\end{figure}

Full-pathway musculoskeletal care is distributed across a long tail of diagnoses, rather than concentrated in a small set of high-volume disorders. OrthoBench captured this breadth with 1{,}000 disease categories from 5{,}905 patients (Fig.~\ref{fig:disease_reader}a). For the radial analysis, we focused on 62 diseases with sufficient case support in the source cohort, comprising 48 non-rare diseases with at least 20 cases and 14 rare diseases with at least 10 cases. These diseases accounted for 3{,}556 patients in the source cohort. To place system-level performance in context, we compared OrthoPilot with four representative systems: CHEESE, its clinical reasoning backbone; GPT-5.1, a frontier proprietary model; DeepSeek-R1, a general reasoning model; and MedGemma-27B, a medical-domain model (Fig.~\ref{fig:disease_reader}b).

Across these 62 diseases, OrthoPilot achieved the highest mean disease-level score among the five systems (75.4\%, compared with 68.9\% for CHEESE, 67.2\% for GPT-5.1, 63.0\% for DeepSeek-R1 and 61.1\% for MedGemma-27B). It ranked first in 40 diseases and exceeded CHEESE, GPT-5.1, DeepSeek-R1 and MedGemma-27B in 43, 50, 57 and 58 diseases, respectively. The leading examples spanned different forms of pathway reasoning. In femoral neck fracture, OrthoPilot reached 91.1\%, a setting in which diagnosis must be linked to operative timing, perioperative risk and discharge planning. In cervical disc herniation and lumbar spinal stenosis, it scored 84.7\% and 70.8\%, respectively. These spine disorders require symptoms, imaging findings and functional limitation to be reconciled before conservative or surgical management is selected. In avascular necrosis of the femoral head, OrthoPilot achieved 86.5\%, consistent with the need to connect disease stage with joint-preserving or reconstructive treatment.

The rare and oncological categories represented a different evidence pattern. Decisions often depend on disease-specific features, reconstruction planning and multidisciplinary coordination. Among the 14 rare diseases shown in the radial analysis, OrthoPilot again had the highest mean score (72.2\%, compared with 64.2\% for GPT-5.1, 61.6\% for CHEESE, 60.4\% for DeepSeek-R1 and 58.9\% for MedGemma-27B). It performed strongly in iliac bone malignancy (87.4\%), lumbar vertebral malignancy (86.0\%), lower-limb soft-tissue malignancy (80.4\%) and hereditary multiple osteochondroma (77.3\%). These findings indicate that OrthoPilot preserved pathway-level reasoning across changing disease contexts, from common trauma and degenerative disease to rare oncological and developmental disorders.

\subsubsection{Reader study on OrthoPilot as an assistant}

We next tested OrthoPilot as both a standalone system and an assistive system during full-pathway case interpretation. We designed a seven-arm washout-crossover reader study using 307 patients from the Ruijin Hospital longitudinal cohort from 2018 to 2023. The cases were stratified by seven disease categories and evaluated by 81 physicians in three experience tiers: Experts ($\geq$25 years, $n = 27$), Specialists (10--24 years, $n = 27$) and Residents (1--9 years, $n = 27$). In the first phase, each physician completed all 11 pathway tasks without AI assistance. After a one-month washout period, the same physicians re-evaluated the same cases with OrthoPilot output available as a reference. OrthoPilot was also evaluated as a standalone seventh arm. The study generated 550{,}451 evaluations, comprising closed-ended diagnostic tasks scored by accuracy and open-ended management tasks scored with ORACLE.

The standalone system exceeded the unassisted physician arms in the complete reader-study table (Supplementary Table~\ref{tab:reader_per_phy}). OrthoPilot achieved a mean overall score of 80.0\%, compared with 65.4\% for Experts, 59.5\% for Specialists and 56.2\% for Residents. This corresponded to gains of 14.6\%, 20.5\% and 23.8\%, respectively. The same pattern was apparent in the task-family summary shown in Fig.~\ref{fig:disease_reader}c. For closed-ended diagnosis, OrthoPilot reached a mean accuracy of 76.4\%, whereas Experts, Specialists and Residents reached 60.9\%, 55.6\% and 52.5\%. For open-ended management, OrthoPilot scored 82.1\% compared with 68.0\%, 61.7\% and 58.2\%. Thus, the system not only improved answer selection. It also produced stronger management responses when physicians had to convert evidence into clinical action.

Providing OrthoPilot output to physicians improved performance in every experience tier. Experts increased from 65.4\% to 76.0\%, Specialists from 59.5\% to 74.6\%, and Residents from 56.2\% to 72.7\%. The absolute gains were largest in Specialists and Residents, indicating that the system provided the greatest benefit where baseline performance left the most room for improvement. After assistance, the three physician tiers occupied a narrow performance interval of 72.7--76.0\%. By contrast, the unaided interval spanned 56.2--65.4\%. AI-assisted Residents also exceeded unassisted Experts by 7.3\%. This pattern is consistent with compression of the experience gap, in which decision support narrowed the difference between junior and senior physicians.

The per-task distributions supported this interpretation (Fig.~\ref{fig:disease_reader}d,e). For the four diagnostic tasks, physician-alone mean accuracy ranged from 52.5\% to 60.9\% across tiers. With OrthoPilot assistance, the range shifted upward to 65.0--68.0\%. The assisted distributions showed closer overlap between tiers, suggesting that the gain was accompanied by reduced inter-physician variability. The paired analysis of the seven open-ended management tasks showed a similar upward shift. Improvements were observed for perioperative assessment, surgical planning, preoperative orders, postoperative orders, discharge summary, rehabilitation planning and multidisciplinary consultation. The complete task-level values for all physician tiers and assisted conditions are provided in Supplementary Table~\ref{tab:reader_per_phy}. These findings indicate that OrthoPilot functioned both as a standalone clinical reasoning system and as an assistive system that raised physician performance across the full musculoskeletal care pathway.

\subsection{Prospective clinical study}
\subsubsection{Clinical decision-making across key musculoskeletal tasks}

\FloatBarrier

\begin{figure}[htbp]
\centering
\includegraphics[width=0.95\textwidth,height=0.85\textheight,keepaspectratio]{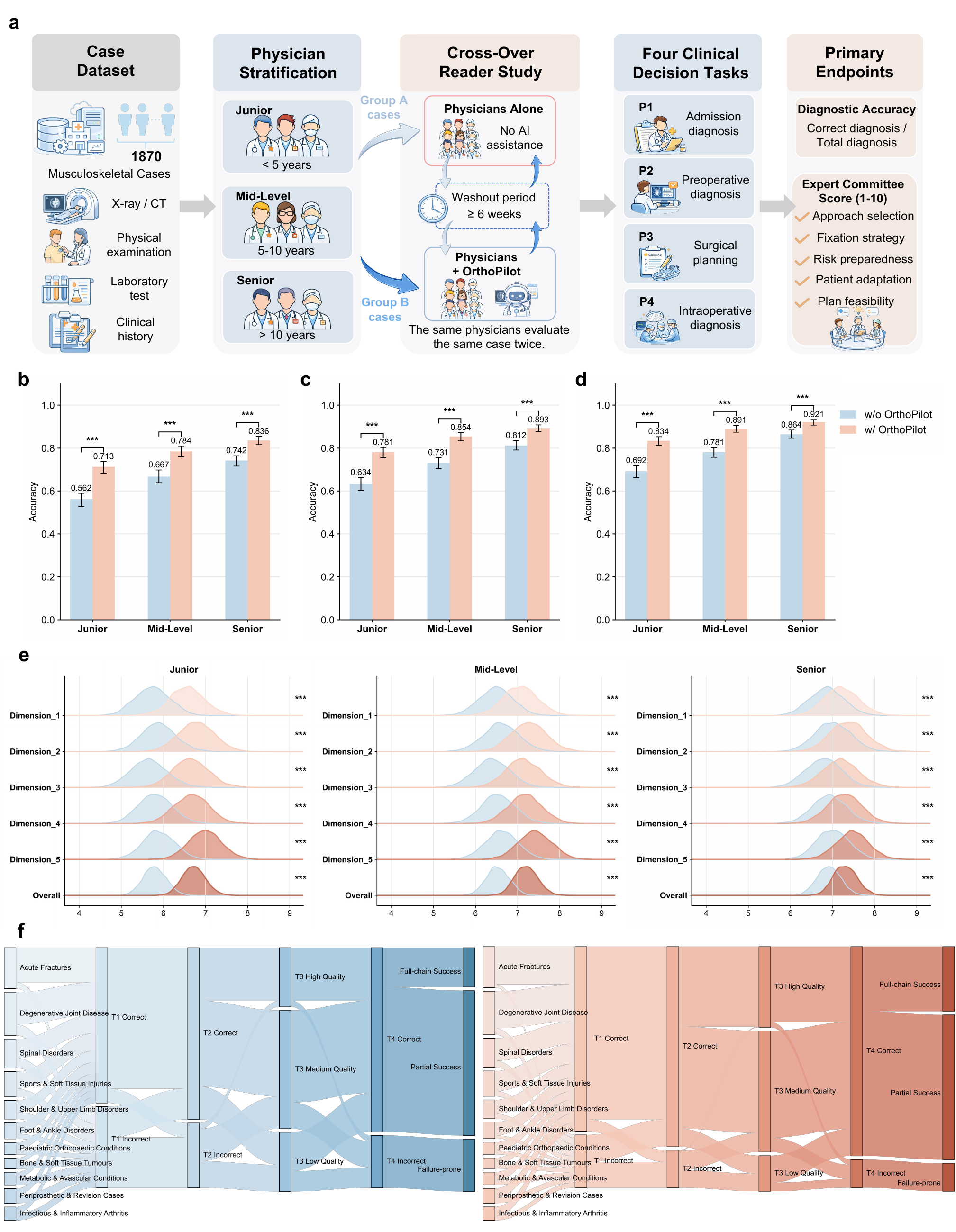}
\caption{\textbf{Prospective crossover reader study evaluating OrthoPilot in musculoskeletal clinical decision-making.}
\textbf{a,} Study design of the prospective crossover reader study. A total of 1,870 cases were evaluated by 30 physicians across three experience tiers, with each tier independently assessing all 1,870 cases, under unassisted and AI-assisted conditions. Four clinical tasks were assessed, including admission diagnosis (P1), preoperative diagnosis (P2), surgical planning (P3) and intraoperative diagnosis (P4).
\textbf{b,c,d,} Case-level diagnostic accuracy for P1, P2 and P4 stratified by physician experience level under unassisted and AI-assisted settings. Error bars indicate 95\% confidence intervals. Statistical significance was assessed using two-sided McNemar tests on paired case-level outcomes. $^{*}P < 0.05,\ ^{**}P < 0.01,\ ^{***}P < 0.001$.
\textbf{e,} Expert committee scores for P3 across five evaluation dimensions and the overall score. Statistical significance was assessed using paired Wilcoxon signed-rank tests.
\textbf{f,} Sequential workflow analysis across P1--P4. P3 quality was stratified into low, medium and high groups according to expert scores, using thresholds derived from the unassisted condition and applied to both conditions.}

\label{fig:crossover reader study}
\end{figure}

To determine whether OrthoPilot improves critical musculoskeletal decision-making, we conducted a prospective crossover reader study using 1,870 complex musculoskeletal cases from Ruijin Hospital, Shanghai, China. Each case comprised multimodal clinical data, including imaging (X-ray, CT or MRI), physical examination findings, laboratory tests and clinical history (Fig.~\ref{fig:crossover reader study}a). Thirty orthopaedic physicians were enrolled and stratified into three experience levels: junior ($<5$ years), mid-level ($5$--$10$ years) and senior ($>10$ years).
On their assigned cases, each physician completed four key clinical decision tasks spanning the core stages of the musculoskeletal workflow: admission diagnosis (P1), preoperative diagnosis (P2), surgical planning (P3) and intraoperative diagnosis (P4).
All eligible cases were randomly divided into two sets and allocated across physicians within each experience stratum (Supplementary Fig.~\ref{fig:ext_prospective_data_flow}). In the first round, each physician evaluated one set without AI assistance and the other with OrthoPilot assistance; after a washout period of at least 6 weeks to reduce recall bias, the same physicians reassessed the identical cases under the opposite condition. This crossover design ensured that every assigned case was assessed once with and once without OrthoPilot assistance by the same physician, yielding paired case-level observations within each experience stratum. Diagnostic performance for P1, P2 and P4 was quantified as accuracy against reference standards (Supplementary Table~\ref{tab:prospective_cohort_reader_study}); case-level accuracy was reported with 95\% confidence intervals, and paired binary outcomes between the unassisted and AI-assisted conditions were compared using two-sided McNemar tests. Surgical planning quality (P3) was rated by a blinded expert committee of fifteen senior orthopaedic surgeons, each with more than 15 years of clinical experience, with every plan independently scored by at least three experts. The final score was calculated as a weighted sum of five predefined dimensions, comprising approach selection, fixation strategy, risk preparedness, patient-specific adaptation and plan feasibility, each rated from 1 to 10 according to prespecified weights (Supplementary Tables~\ref{tab:surgical_planning_criteria} and \ref{tab:score_interpretation}). For the full-pathway analysis, surgical planning quality was categorized as low, medium or high using quartile thresholds derived from the distribution of case-level mean expert scores under the unassisted condition; the same thresholds were applied to both conditions, classifying scores below the first quartile as low quality, those between the first and third quartiles as medium quality, and those above the third quartile as high quality.

Across all four tasks and physician experience levels, OrthoPilot assistance consistently improved clinical decision performance. Among the diagnostic tasks, the largest gain was observed for admission diagnosis, where overall P1 accuracy rose from $0.657$ to $0.778$, with particularly marked improvements in the most challenging disease categories (Fig.~\ref{fig:crossover reader study}b). Subgroup analyses further showed that the benefit was most pronounced among junior physicians. In P1, for example, case-level accuracy among junior physicians improved from $0.562$ (95\% CI $0.528$--$0.589$) to $0.713$ (95\% CI $0.683$--$0.737$; two-sided McNemar test, $P < 0.001$), with similar trends for preoperative and intraoperative diagnosis (Fig.~\ref{fig:crossover reader study}c,d). These results indicate that OrthoPilot improved overall diagnostic accuracy while narrowing performance gaps across physician experience levels.
For surgical planning, OrthoPilot significantly improved blinded expert committee scores across all five evaluation dimensions, namely approach selection, fixation strategy, risk preparedness, patient-specific adaptation and plan feasibility. Among junior physicians, the overall score increased from $5.794$ (95\% CI $5.785$--$5.802$) to $6.721$ (95\% CI $6.717$--$6.728$; $P < 0.001$), with corresponding gains also observed in mid-level and senior physicians (Fig.~\ref{fig:crossover reader study}e).
At the distribution level, applying thresholds derived from the unassisted condition, OrthoPilot shifted surgical planning quality toward more favorable categories, raising the proportion of cases classified as high quality from $25.0\%$ to $34.0\%$ while lowering the proportion classified as low quality from $25.0\%$ to $15.0\%$ (Fig.~\ref{fig:crossover reader study}f).
More importantly, these task-level improvements translated into substantial gains across the full clinical workflow. The proportion of cases achieving full-chain success increased from $310$ of $1{,}870$ ($16.6\%$) to $509$ of $1{,}870$ ($27.2\%$), whereas failure-prone outcomes decreased from $413$ cases ($22.1\%$) to $221$ cases ($11.8\%$). Together, these findings demonstrate that OrthoPilot improved not only single-step decision accuracy but also the overall robustness of sequential musculoskeletal decision-making, increasing successful pathways while reducing clinically vulnerable ones.

\subsubsection{System-wide clinical impact across the musculoskeletal care continuum}

\FloatBarrier

\begin{figure}[htbp]
\centering
\includegraphics[width=\textwidth,height=0.85\textheight,keepaspectratio]{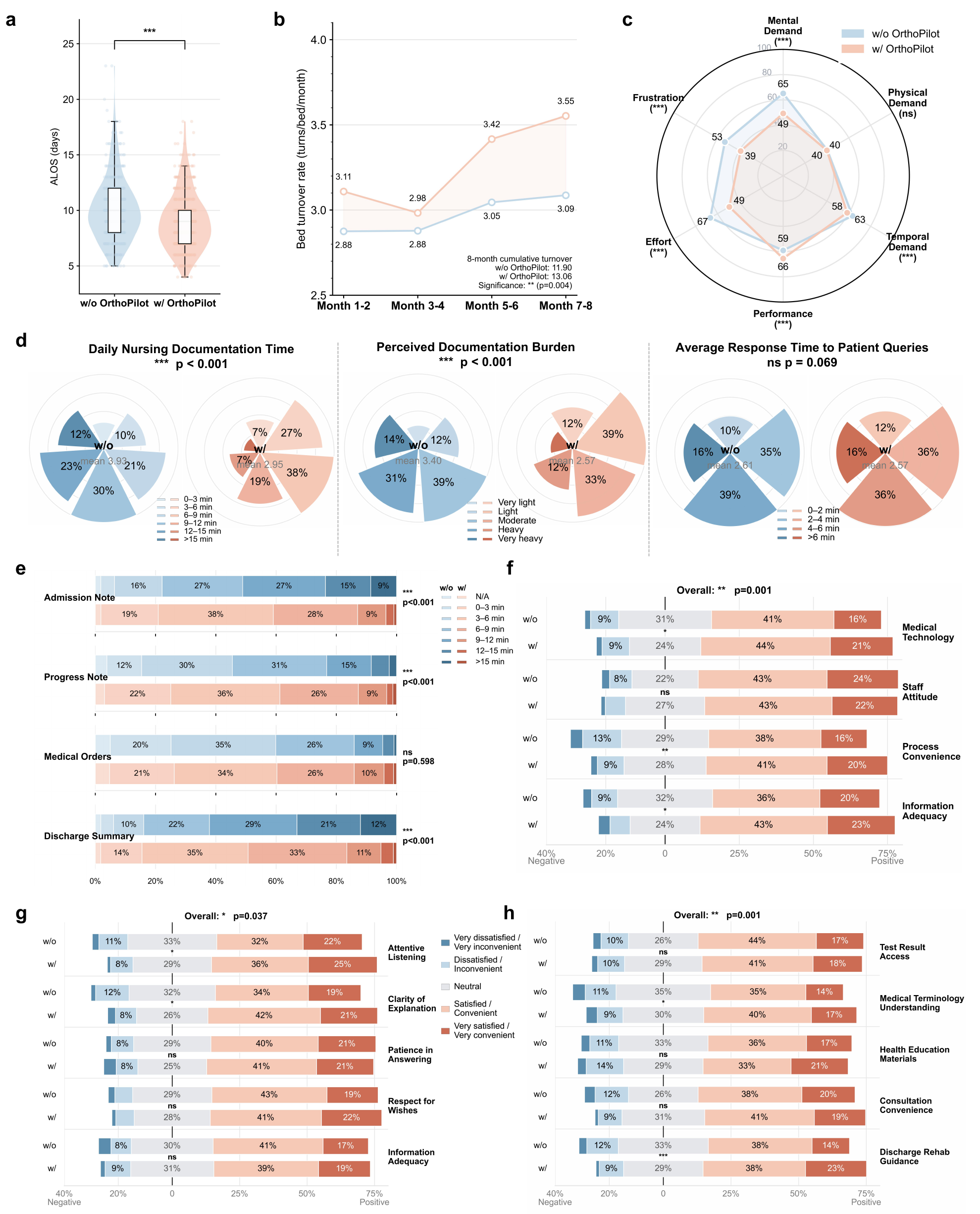}
\caption{\textbf{Impact of OrthoPilot on clinical efficiency, workload and patient experience across the musculoskeletal care continuum.}
\textbf{a,b,} Hospital-level efficiency metrics, including ALOS and cumulative bed turnover over an 8-month period.
\textbf{c,} Physician workload evaluated using NASA-TLX dimensions.
\textbf{d,} Nursing workflow evaluation, including documentation time, documentation burden and response time to patient queries.
\textbf{e,} Physician documentation efficiency across admission notes, progress notes, discharge summaries and medical order entry.
\textbf{f,g,h,} Patient-reported outcomes, including overall satisfaction, communication quality and accessibility of health information.
Statistical significance was assessed using two-sided t-tests (\textbf{a,b}), paired Wilcoxon signed-rank tests (\textbf{c,d}), chi-squared tests (\textbf{e}) and two-sided Mann--Whitney U tests (\textbf{f,g,h}). *$P < 0.05$, **$P < 0.01$, ***$P < 0.001$, ns, not significant.}

\label{fig:Pilot Deployment}    
\end{figure}

To evaluate the real-world clinical impact of OrthoPilot, we conducted a single-center, prospective randomized controlled study at Ruijin Hospital, Shanghai, China, over an 8-month period from September 2025 to April 2026. A total of 8,240 consecutive musculoskeletal inpatients admitted for diagnostic assessment or surgical treatment were enrolled and randomly assigned in a 1:1 ratio to either the standard-care group or the OrthoPilot-assisted group (Supplementary Fig.~\ref{fig:ext_prospective_data_flow}). Randomization was stratified by major disease category, surgical complexity and admission type to improve balance between groups. In the control group, patients were managed according to the conventional clinical workflow without OrthoPilot. In the intervention group, OrthoPilot was integrated into the hospital information system and provided structured decision support before final clinical decision-making, while all final decisions remained under physician responsibility. OrthoPilot operated on multimodal clinical data, including electronic medical records, radiographic imaging, laboratory results and real-time vital signs, and supported the full musculoskeletal care pathway, spanning pre-diagnostic evaluation, surgical decision and execution, postoperative management, and rehabilitation planning.
The clinical impact of OrthoPilot was prospectively evaluated across four predefined domains, including hospital-level efficiency, physician workload, nursing workflow and patient-reported experience. Hospital-level endpoints, including average length of stay (ALOS) and bed turnover rate, were extracted from hospital administrative records. Physician and nursing outcomes, including documentation time and workload, were collected using structured logs and validated instruments such as the NASA-TLX scale. Patient-reported outcomes, including satisfaction, communication quality and information accessibility, were collected using standardized Likert-scale questionnaires.

Across all evaluation domains, integration of OrthoPilot consistently enhanced clinical efficiency, reduced workload, and improved patient experience. At the hospital scale, OrthoPilot was associated with a significant reduction in ALOS ($P < 0.001$) and a substantial increase in cumulative bed turnover over the 8-month study period, rising from 11.90 to 13.06 cases per bed ($P = 0.004$), indicating improved resource utilization and operational efficiency (Fig.~\ref{fig:Pilot Deployment}a,b).
Among physicians, OrthoPilot markedly alleviated cognitive workload, with reductions in mental demand (from 65 to 49), effort (from 67 to 49), frustration (from 53 to 39), and temporal demand (from 63 to 58), alongside an increase in perceived performance (from 59 to 66) (all $P < 0.001$), while physical demand remained unchanged (Fig.~\ref{fig:Pilot Deployment}c).
Parallel improvements were observed in nursing workflows, where OrthoPilot reduced both daily documentation time (mean score from 3.93 to 2.95, $P < 0.001$) and perceived documentation burden (mean score from 3.40 to 2.57, $P < 0.001$), accompanied by a marked decline in high-duration documentation tasks. In contrast, average response time to patient queries remained unchanged (mean score from 2.61 to 2.57, $P = 0.069$) (Fig.~\ref{fig:Pilot Deployment}d).
Documentation efficiency was also improved, as reflected by a substantial decrease in time spent in longer-duration categories ($>9$ min) for admission notes, progress notes, and discharge summaries (all $P < 0.001$), whereas structured medical order entry showed no significant change ($P = 0.598$) (Fig.~\ref{fig:Pilot Deployment}e).
From the patient perspective, OrthoPilot significantly improved overall satisfaction ($P = 0.001$), communication quality ($P = 0.037$), and accessibility of health information ($P = 0.001$). These gains were primarily driven by information-related dimensions, where the proportion of patients reporting positive experiences increased significantly in medical technology (from 57\% to 65\%, $P < 0.05$), process convenience (from 54\% to 61\%, $P < 0.01$), and information adequacy (from 56\% to 66\%, $P < 0.05$). Furthermore, improvements were noted in the clarity of explanation (from 53\% to 63\%, $P < 0.05$), medical terminology understanding (from 49\% to 57\%, $P < 0.05$), and discharge rehabilitation guidance (from 52\% to 61\%, $P < 0.001$). In contrast, interpersonal aspects of care, including staff attitude, patience in answering, and respect for wishes, showed no significant differences between groups (Fig.~\ref{fig:Pilot Deployment}f--h).
Taken together, these findings indicate that OrthoPilot enhances system-level performance across the musculoskeletal care continuum, improving efficiency and information delivery while preserving the human-centered aspects of care.

\FloatBarrier

\section{Discussion}\label{sec12}

This study establishes OrthoPilot as a clinically validated agentic AI system for full-pathway musculoskeletal management. It addresses a central limitation of current clinical AI: most systems are designed to answer isolated questions, whereas real care depends on decisions that unfold across a changing patient trajectory. Musculoskeletal disease makes this challenge explicit. Management is not a single act after diagnosis, but a sequence of decisions that must protect function as patient evidence, therapeutic goals and recovery needs evolve. OrthoPilot was built to meet this higher bar. Its dual-evidence architecture links patient-specific evidence from hospital systems with external medical knowledge. A CHEESE-driven agent maintains the clinical objective and pathway state, identifies missing evidence and uses Tool Plaza to retrieve focused information before generating traceable recommendations. This design brings patient records, clinical tools and external knowledge into one auditable reasoning process across admission, surgery, discharge and rehabilitation. Our findings therefore move clinical AI beyond model-centric competence towards evidence-grounded clinical agency, in which the system acquires the evidence needed at each point in care and supports longitudinal management across the complete pathway.

The study highlights a persistent gap between model competence and clinical utility. Recent medical AI systems have shown important progress in expert-level question answering and multimodal clinical reasoning \cite{Singhal2025MedPaLM2,Saab2025MedGemini}. Collaborative agent-based decision-making has also begun to emerge \cite{Breazeal2024MDAgents}. These studies show that language models can acquire substantial medical knowledge and apply it to demanding clinical tasks. Existing evidence, however, remains centred on bounded prompts, diagnostic dialogues or predefined benchmark items. Clinical decisions occur along an evolving patient trajectory. Evidence may be missing, distributed across systems or altered by new imaging, operative findings or rehabilitation goals. To capture this structure, our evaluation framework assessed full-pathway musculoskeletal management and scored open-ended recommendations against physician-defined decision elements rather than lexical overlap, as detailed in Methods and Appendix Section~\ref{sec:evaluation_framework_benchmark}. This framing extends the evaluation of clinical agents beyond single-time-point performance. Clinically useful agents need to preserve patient state, update evidence after clinical action and align subsequent recommendations with the evolving trajectory of care.

The evaluation programme showed that OrthoPilot remained useful across increasingly demanding clinical settings. Rather than treating scale as an endpoint, we used scale to test continuity, generalisation and deployability. The 21-year longitudinal cohort of 183,472 patients, together with external validation across 60 centres, provided the clinical substrate for modelling musculoskeletal care as a pathway. Ablation analyses then separated the sources of this behaviour. Performance gains were preserved when the workflow was fixed across backbone models, whereas full-pathway reasoning declined when evidence tools were removed, as detailed in Supplementary Section~\ref{sec:ablation_study}. This indicates that the relevant mechanism was the coupling of language-model reasoning to patient-grounded evidence, external knowledge access and pathway orchestration. Reader studies tested whether this mechanism improved human decisions. More than half a million physician-scored decisions showed that assistance improved decisions while reducing experience-dependent variation. These findings extend prior observations that technical performance alone does not establish clinical usefulness \cite{Sutton2020CDSS,Mittermaier2023DecisionSupport,Han2024AIRCT}. A practical test for clinical agents is whether they preserve evidence across decisions that are separated in time. A recommendation is useful only if it remains coherent with the evolving record and can be reviewed by clinicians.

Prospective validation tested whether this design could survive contact with routine care. This question matters because health systems worldwide increasingly divide care across admission, surgery, discharge and rehabilitation, leaving clinicians to reconnect evidence that appears at different times. In the physician decision-making study, assistance increased success across the complete management chain rather than at only one decision point. In the 8,240-patient inpatient deployment, OrthoPilot was embedded in hospital workflows and supported decisions as the patient record changed. This is different from a system that returns a static suggestion, such as requesting another test, while leaving the pathway unchanged. OrthoPilot can retrieve new imaging and laboratory evidence, reassess perioperative readiness, propose medication and monitoring orders, revise discharge documentation and generate rehabilitation plans for clinician review. These results suggest that agentic clinical AI should be judged by its ability to operate within governed clinical work, not only by performance on predefined prompts. This clinical embedding also clarifies the governance problem. A locally deployed, language-model-powered system can expose the evidence it retrieves, the tools it accesses and the rationale shown to users. Clinicians can therefore review not only the final recommendation, but also how patient evidence was converted into an action. This is important because the system is not intended to store all medical knowledge in model parameters. Its purpose is to acquire relevant evidence at run time and translate it into patient-specific next steps. Musculoskeletal care provides a stringent test of this design because evidence appears at different points in the admission and must be converted into different actions, from perioperative orders to rehabilitation planning. Similar longitudinal problems arise in oncology, critical care and surgical care, where clinical state changes after treatment and monitoring \cite{Tomasev2019AKI,Hashimoto2018AISurgery}. The transferable contribution is therefore not a fixed orthopaedic workflow, but a governed design principle: connect changing patient evidence to concrete clinical action under clinician review. To our knowledge, this is the first large-scale prospective deployment of an agentic clinical AI system for management across a full musculoskeletal care pathway.

Despite these promising early results, several limitations and concerns warrant consideration. OrthoPilot reasons over an evidence landscape that is inherently uneven. External knowledge sources differ in timeliness, coverage and local relevance. Hospital data also reflect the documentation practices of the institutions that generate them. Some clinically important information may remain outside the retrievable evidence space, including local protocols, unpublished experience or sources that are not openly accessible. Retrieval may therefore omit relevant knowledge or propagate uncertain evidence into downstream recommendations. A key future direction is enhanced provenance. Claims could be made traceable not only to a source, but to the specific passage, figure, table or patient-record element that supports them. This would help clinicians judge when a recommendation is grounded in reliable evidence and when it reflects a weak or conflicting source. OrthoPilot also inherits the risk of language-model hallucination, especially when evidence is sparse, outdated or contradictory \cite{Hager2024LLMLimitations,Gu2024ProbabilisticPredictions}. The present studies also do not yet establish whether improved workflow continuity leads to durable clinical benefit after discharge.

Privacy and ethical concerns also become more salient as clinical agents move closer to hospital information systems. A full-pathway agent may require access to admission notes, imaging reports, laboratory results, operating-room records, nursing documentation and rehabilitation plans. These data are sensitive and are not equally relevant to every user. Implementation could therefore make access contingent on role, clinical purpose and institutional policy. Physicians, nurses, rehabilitation teams and patients may need different views of the same longitudinal record. Access could be logged, reviewable and limited to what is needed for care. Such safeguards are not separate from model performance. They define whether the use of patient evidence is necessary, visible and accountable during clinical management.

To extend OrthoPilot towards routine full-pathway clinical management, future work could aim to strengthen four aspects of the system. A near-term direction is to improve robustness by broadening access to reliable external knowledge, checking retrieved evidence more rigorously and linking each recommendation more clearly to the supporting patient record. It may also be useful to connect OrthoPilot more deeply with hospital information systems. This could expand the tool space available to the agent, including imaging systems, laboratory systems, order entry, rehabilitation scheduling and follow-up tools. Such integration would allow management recommendations to be generated within the clinical systems where they are reviewed and acted on. Broader multicentre deployment could then provide more stringent tests of reliability. Hospitals differ in information systems, documentation habits and governance rules, and these differences would help define the boundaries of generalisation. Another promising avenue is patient-level memory. Long-term memory could preserve relevant history, prior recommendations, clinician revisions and recovery goals across encounters, supporting more personalised management. In time, useful clinical trajectories could be distilled into reusable skills. Aggregated and audited trajectories may also help update the external knowledge world, including local protocols and guideline-like resources. This would move OrthoPilot towards a clinician-governed loop between patient evidence, management action and clinical knowledge. In conclusion, we developed and validated a language-model-powered clinical agent for full-pathway musculoskeletal management. This work establishes a clinically tested route towards AI-supported management across the complete care pathway.

\bibliographystyle{unsrtnat}
\bibliography{references}

\newpage

\section{Methods}\label{sec:methods}

\subsection{Problem Formulation}

\subsubsection{Full-pathway task definition}

We formulate musculoskeletal care reasoning as a longitudinal sequence of clinically grounded evaluation nodes distributed across the inpatient care pathway, rather than as a collection of isolated prediction tasks. This formulation follows the logic of real musculoskeletal care, in which clinical decisions are generated progressively through evidence accumulation, procedural intervention, discharge transition, and post-discharge recovery. Importantly, the pathway is not strictly linear: while most nodes are temporally ordered, the preoperative preparation phase contains an iterative decision loop in which order planning and operability assessment may recur until surgical readiness is established.

Formally, for a patient trajectory $\mathcal{P}$, we denote the full-pathway reasoning process as an ordered sequence of stage-specific states:
\begin{equation}
\mathcal{P} = (s_1, s_2, \ldots, s_{11}),
\end{equation}
where each $s_t$ denotes the patient state at a key evaluation node. At each node, the system generates a task-conditioned clinical output:
\begin{equation}
y_t = f_{\theta}(s_t), \qquad t = 1,\ldots,11,
\end{equation}
where $f_{\theta}$ denotes the LLM-based reasoning function and $y_t$ denotes the structured output associated with node $t$. Because the pathway evolves through sequential evidence acquisition and intervention, the patient state is updated recursively as:
\begin{equation}
s_t = \Phi(s_{t-1}, \Delta_t),
\end{equation}
where $\Delta_t$ denotes newly acquired evidence, observations, or interventions introduced at stage $t$, and $\Phi(\cdot)$ denotes the state-transition operator over the clinical trajectory.

Under this formulation, the pathway begins with admission diagnosis (Task~1; $\mathcal{N}_{\mathrm{AdmDx}}$), which requires the system to transform the patient's initial presentation into a preliminary diagnostic hypothesis under substantial uncertainty. The pathway then enters a preoperative preparation phase centred on two coupled nodes. The first is preoperative orders (Task~7; $\mathcal{N}_{\mathrm{PreopOrder}}$), in which the system proposes the next-step work-up and perioperative preparation required to advance clinical decision-making. The second is perioperative assessment (Task~5; $\mathcal{N}_{\mathrm{PeriAssess}}$), in which the system evaluates whether surgery can proceed on the basis of accumulated imaging, laboratory, cardiovascular, infectious, and other perioperative evidence. In practice, these two nodes form a local evidence-acquisition loop: if current evidence is insufficient to determine surgical readiness, the system must first issue additional preoperative orders, retrieve the corresponding results, and then re-evaluate perioperative readiness.

Once operative readiness has been established, the pathway advances to preoperative diagnosis (Task~2; $\mathcal{N}_{\mathrm{PreopDx}}$), where the system consolidates all presurgical evidence into a definitive preoperative diagnosis and, where appropriate, a subtype or anatomical characterization. Multidisciplinary consultation (Task~11; $\mathcal{N}_{\mathrm{MDT}}$) is defined as a cross-stage node that may be invoked before or after surgery. It evaluates whether the system can retrieve specialist recommendations from relevant departments and integrate cross-disciplinary opinions into the evolving care plan, including preoperative risk stratification and optimization, postoperative complication management, rehabilitation planning, and discharge coordination. On this basis, surgical planning (Task~6; $\mathcal{N}_{\mathrm{SurgPlan}}$) translates the consolidated diagnostic and multidisciplinary evidence into a concrete operative plan, including approach selection, implant specification, and contingency considerations. After intervention, Intraoperative diagnosis (Task~3; $\mathcal{N}_{\mathrm{IntraopDx}}$) tests whether the system can update its diagnostic understanding in light of intraoperative findings and postoperative information. Postoperative orders (Task~8; $\mathcal{N}_{\mathrm{PostopOrder}}$) then evaluate immediate postoperative management, including analgesia, thromboprophylaxis, monitoring, imaging or laboratory follow-up, and early rehabilitation instructions.

The later nodes assess whether the model can sustain coherent reasoning beyond acute intervention. Discharge diagnosis (Task~4; $\mathcal{N}_{\mathrm{DischDx}}$) examines whether the system can synthesize the full hospital course into a final account of the principal condition and its clinically relevant comorbidities. Discharge summary (Task~9; $\mathcal{N}_{\mathrm{DischSum}}$) evaluates whether the admission course and major interventions can be compressed into a clinically coherent transition document. Rehabilitation planning (Task~10; $\mathcal{N}_{\mathrm{Rehab}}$) tests whether the system can formulate staged recovery goals, home exercise instructions, and outpatient follow-up plans, potentially incorporating post-discharge clinic records when available.

The mapping between benchmark task numbers and pathway evaluation nodes is summarised in Table~\ref{tab:task_node_mapping}. Tasks~1--4 correspond to the four diagnostic reasoning nodes and are evaluated using closed-ended accuracy, whereas Tasks~5--11 correspond to the seven clinical management nodes and are evaluated using the ORACLE framework.

\begin{table}[htbp]
\centering
\caption{\textbf{Correspondence between benchmark task numbers and care-pathway evaluation nodes.} Tasks~1--4 are closed-ended diagnostic reasoning tasks; Tasks~5--11 are open-ended clinical management tasks. The pathway column indicates the temporal position along the musculoskeletal care trajectory.}
\label{tab:task_node_mapping}
\small
\begin{tabular}{@{}clll@{}}
\toprule
Task & Clinical task name & Node symbol & Pathway position \\
\midrule
\multicolumn{4}{@{}l}{\textit{Diagnostic reasoning tasks}} \\
1 & Admission diagnosis & $\mathcal{N}_{\mathrm{AdmDx}}$ & Admission \\
2 & Preoperative diagnosis & $\mathcal{N}_{\mathrm{PreopDx}}$ & Pre-surgery \\
3 & Intraoperative diagnosis & $\mathcal{N}_{\mathrm{IntraopDx}}$ & Intraoperative \\
4 & Discharge diagnosis & $\mathcal{N}_{\mathrm{DischDx}}$ & Discharge \\
\midrule
\multicolumn{4}{@{}l}{\textit{Clinical management tasks}} \\
5 & Perioperative assessment & $\mathcal{N}_{\mathrm{PeriAssess}}$ & Preoperative loop \\
6 & Surgical planning & $\mathcal{N}_{\mathrm{SurgPlan}}$ & Pre-surgery \\
7 & Preoperative orders & $\mathcal{N}_{\mathrm{PreopOrder}}$ & Preoperative loop \\
8 & Postoperative orders & $\mathcal{N}_{\mathrm{PostopOrder}}$ & Post-surgery \\
9 & Discharge summary & $\mathcal{N}_{\mathrm{DischSum}}$ & Discharge \\
10 & Rehabilitation planning & $\mathcal{N}_{\mathrm{Rehab}}$ & Post-discharge \\
11 & Multidisciplinary consultation & $\mathcal{N}_{\mathrm{MDT}}$ & Pre- and post-surgery \\
\bottomrule
\end{tabular}
\end{table}

The full-pathway task sequence can therefore be written as:

\begin{equation}
\begin{aligned}
\mathcal{T}_{\mathrm{path}} = \big(&
\mathcal{N}_{\mathrm{AdmDx}},\;
(\mathcal{N}_{\mathrm{PreopOrder}} \leftrightarrow \mathcal{N}_{\mathrm{PeriAssess}})^{*},\;
\\[-0.15em]
&
\mathcal{N}_{\mathrm{PreopDx}},\;
[\mathcal{N}_{\mathrm{MDT}}]_{\mathrm{pre}},\;
\mathcal{N}_{\mathrm{SurgPlan}},\;
\\[-0.15em]
&
\mathcal{N}_{\mathrm{IntraopDx}},\;
\mathcal{N}_{\mathrm{PostopOrder}},\;
[\mathcal{N}_{\mathrm{MDT}}]_{\mathrm{post}},\;
\\[-0.15em]
&
\mathcal{N}_{\mathrm{DischDx}},\;
\mathcal{N}_{\mathrm{DischSum}},\;
\mathcal{N}_{\mathrm{Rehab}}
\big),
\end{aligned}
\end{equation}
where $(\mathcal{N}_{\mathrm{PreopOrder}} \leftrightarrow \mathcal{N}_{\mathrm{PeriAssess}})^{*}$ denotes the potentially repeated preoperative sub-loop between order planning and perioperative assessment. The bracketed MDT terms denote stage-contingent multidisciplinary consultation before or after surgery when cross-disciplinary input is clinically required; they refer to the same Task~11 node rather than separate benchmark tasks. The remaining elements denote temporally ordered evaluation nodes along the musculoskeletal care pathway. Note that the benchmark task numbers (Tasks~1--11) follow a category-based ordering (diagnostics first, then management), whereas the pathway expression above follows the temporal order of clinical care; Table~\ref{tab:task_node_mapping} provides the explicit mapping between these two orderings. Accordingly, the benchmark measures not only node-level accuracy but also whether reasoning remains stable, context-sensitive, and clinically coherent across both iterative local loops and the overall longitudinal trajectory.

\subsubsection{Task inputs and outputs}

At each evaluation node, the input consists of the patient-specific information available up to that stage of care, including structured and unstructured records from the in-hospital information world and, when required, supplementary evidence retrieved from the external knowledge world. These inputs may include admission notes, history of present illness, demographic information, physical examination findings, radiology reports and images, laboratory measurements, pathology results, operative documentation, prior clinical records, outpatient follow-up notes, and consultation records.
Formally, the stage-specific input is defined as:
\begin{equation}
x_t = (\mathcal{X}_{\mathrm{in}}^{(t)}, \mathcal{K}_{\mathrm{ext}}^{(t)}),
\end{equation}
where $\mathcal{X}_{\mathrm{in}}^{(t)}$ denotes the in-hospital evidence available at stage $t$, and $\mathcal{K}_{\mathrm{ext}}^{(t)}$ denotes the external knowledge retrieved for that stage when needed. The corresponding output is a structured clinical recommendation or document appropriate to that node and is represented as:
\begin{equation}
y_t = f_{\theta}\!\left(\mathcal{X}_{\mathrm{in}}^{(t)}, \mathcal{K}_{\mathrm{ext}}^{(t)}\right).
\end{equation}
Examples include an initial or final diagnosis, a preoperative plan, an operability judgment, an operative strategy, postoperative orders, a discharge document, a rehabilitation plan, or an integrated consultation recommendation.

This formulation treats musculoskeletal care as a sequential reasoning problem in which each decision is conditioned on evidence accumulated from earlier stages. It also captures local iterations, such as repeated preoperative order planning and perioperative assessment before surgical readiness is established. The benchmark uses this pathway structure to define the information available at each node and to evaluate whether the system produces appropriate outputs for the corresponding clinical task.

\subsection{OrthoPilot System}

\subsubsection{System architecture}

To bridge the gap between decontextualized conversational models and the dynamic requirements of real clinical workflows, we developed OrthoPilot, a clinical-grade musculoskeletal care agent system designed for full-pathway care. Rather than treating clinical reasoning as a series of isolated predictions, OrthoPilot is built around the concept of clinical agency, enabling sequential decision-making across a longitudinal, state-dependent musculoskeletal care trajectory.
At the core of OrthoPilot lies the Clinical Holistic Evidence-to-Execution Synthesis Engine (CHEESE), a large language model (LLM) with 32 billion parameters. CHEESE is designed to transform heterogeneous clinical evidence into structured and actionable decisions at successive evaluation nodes along the care pathway. Instead of responding only to information explicitly present in the prompt, it can identify evidence gaps, retrieve patient-specific and external medical evidence through the dual evidence worlds, and synthesize clinically grounded recommendations while preserving reasoning coherence across long-horizon and partially iterative clinical trajectories.

\subsubsection{Dual evidence worlds}

The core methodological framework of OrthoPilot is built upon a dual evidence world paradigm. In implementation, the Tool Plaza exposes in-hospital records and external medical knowledge as typed, OpenAI-compatible tools that the LLM can call during reasoning. A tool registry standardizes each tool as a JSON-based function schema, including its name, description, required parameters, and response structure. During inference, the agent runtime converts selected tool calls into JSON requests to independent FastAPI services, which retrieve evidence and return JSON-formatted structured responses for reinsertion into the reasoning context.

The in-hospital information world returns patient-specific evidence from clinical records, imaging reports, laboratory tests, pathology reports, consultation notes and similar cases. The external knowledge world returns evidence from medical knowledge graphs, literature and textbook retrieval, case-report retrieval and web search. This design separates patient factual anchoring from external evidence retrieval, while presenting both sources to CHEESE through the same tool calling interface.

\paragraph{In-hospital information world.}
The in-hospital information world provides the agent with controlled, simulated access to the hospital information system (HIS). Through this tool layer, the agent can query heterogeneous patient information, including clinical notes, radiology reports, laboratory measurements, pathology records, consultation notes and operative documentation. The inputs to this world are patient identifiers, stage-specific clinical queries, and previously accumulated pathway context; the outputs are structured patient-level evidence that can be directly incorporated into downstream reasoning.

Given a stage-specific patient query $q_t$, the in-hospital evidence returned by the system can be written as:
\begin{equation}
\mathcal{X}_{\mathrm{in}}^{(t)} = \bigcup_{n=1}^{N} \mathcal{X}_{n}(q_t),
\end{equation}
where $\mathcal{X}_{n}(\cdot)$ denotes the $n$-th in-hospital retrieval module and $N$ is the number of in-hospital evidence channels. In practice, this evidence world consists of the following tool families.

\textbf{Clinical record retrieval.}
This tool family retrieves structured and narrative clinical records from the hospital electronic health record (EHR) system. The clinical record environment is derived from the Ruijin Hospital longitudinal orthopaedic cohort of 183{,}472 patients admitted between 2004 and 2024; detailed cohort partitioning is provided in Appendix~\ref{sec:retrospective_dataset_curation}. Cases reserved for the historical similar-case retrieval resource were kept separate and were not processed as benchmark cases in this tool family. Given a patient identifier, the tools return admission-level demographic and administrative fields, admission notes and operative records when available. To prevent answer leakage, diagnosis names corresponding to the target evaluation node are masked before delivery. Record access is also constrained by clinical time, so documents generated after a given evaluation node are not exposed at that node. For example, operative records cannot be accessed before the surgical stage. Personally identifiable information and unnecessary free-text fields are removed at the server side before delivery. These records provide the longitudinal patient context used to anchor subsequent stage-specific evidence retrieval.

\textbf{Radiological imaging retrieval.}
These tools retrieve imaging examination results and associated radiology reports from the hospital radiology information system (RIS) and picture archiving and communication system (PACS). Their function is to expose anatomical, structural, and lesion-level evidence for tasks such as fracture characterization, preoperative templating, and postoperative surveillance. The imaging environment encompasses 679{,}468 imaging studies (an average of 3.7 studies per patient) across 4{,}097 distinct examination codes, covering:
\begin{itemize}
    \item conventional radiography (234{,}024 studies, 34.4\%), including anteroposterior, lateral, and oblique X-ray projections;
    \item computed tomography (117{,}219 studies, 17.3\%), including thin-slice reconstructions and three-dimensional reformats;
    \item magnetic resonance imaging (60{,}530 studies, 8.9\%), including T1-weighted, T2-weighted, and fat-suppressed sequences;
    \item ultrasonography (23{,}444 studies, 3.5\%), including musculoskeletal, vascular, and echocardiographic examinations;
    \item other modalities (244{,}251 studies, 35.9\%), including digital subtraction angiography (DSA), dual-energy X-ray absorptiometry (DEXA), PET-CT, SPECT, and nuclear medicine studies.
\end{itemize}
Each tool invocation specifies a patient identifier and a clinical stage (preoperative or postoperative); the returned evidence consists of modality-specific findings and structured report text suitable for downstream diagnostic and planning reasoning.

\textbf{Laboratory diagnostics retrieval.}
These tools retrieve longitudinal laboratory measurements from the hospital information environment. Their role is to provide time-stamped physiological and biochemical evidence for monitoring disease evolution, perioperative readiness, infection risk, metabolic status, and postoperative surveillance. The laboratory environment covers 16.31 million laboratory tests (an average of 88.9 tests per patient) across 2{,}189 test item types, organized into:
\begin{itemize}
    \item haematologic panels, including complete blood count, coagulation profile (PT, APTT, INR), erythrocyte sedimentation rate, and blood typing;
    \item biochemical panels, including hepatic function (ALT, AST, ALP, bilirubin), renal function (creatinine, urea, eGFR), electrolytes (potassium, sodium, chloride, calcium, phosphorus), glucose metabolism, and lipid profiles;
    \item immunological and inflammatory markers, including C-reactive protein, procalcitonin, tumour markers, and autoimmune antibody panels.
\end{itemize}
The tools accept a patient identifier and clinical stage, returning structured laboratory results with reference ranges and abnormality flags, enabling the agent to identify clinically significant deviations and temporal trends across the perioperative timeline.

\textbf{Pathological analysis retrieval.}
These tools access pathology reports and related microscopic evidence from hospital pathology records. Their role is to expose cell- and tissue-level evidence for cases in which gross clinical and imaging presentation is insufficient, especially in oncological, inflammatory, or histologically heterogeneous orthopaedic disorders. The pathology environment covers 19{,}126 patients and 34{,}095 pathology events, organized into:
\begin{itemize}
    \item histopathology, including routine tissue sections and surgical specimen analysis;
    \item cytopathology, including fine-needle aspiration cytology and exfoliative cytology;
    \item molecular pathology, including immunohistochemistry, in situ hybridisation and other molecular assays;
    \item cytogenetics, including chromosome-level and gene-level abnormality assessment;
    \item diagnosis-based pathology categories, which group pathology evidence by clinically assigned diagnostic labels.
\end{itemize}
Given a patient identifier and clinical stage, the tools return pathological findings that refine diagnosis, tumour staging, and subsequent treatment planning.

\textbf{Physician order retrieval.}
These tools retrieve structured physician orders from HIS, comprising 8,021,427 order records spanning 6{,}457 distinct order types. Their role is to expose the temporal sequence of clinical decisions as encoded in order records, enabling the agent to reconstruct the treating physician's diagnostic and therapeutic intent. Each order record contains the order content, order name, order type, signature time, start time, stop time, order status, dosage, dosage unit, quantity, administration route and frequency. The order environment is organized into four categories:
\begin{itemize}
    \item medication orders, including drug name, dosage, administration route, frequency and duration;
    \item diagnostic orders, encompassing requests for laboratory tests, imaging studies and pathological examinations;
    \item nursing orders, including vital sign monitoring schedules, wound care protocols, activity restrictions and fall prevention measures;
    \item therapeutic orders, including physical therapy, rehabilitation programmes and pain management regimens.
\end{itemize}
By exposing the complete order timeline, these tools allow the agent to infer the clinical rationale underlying prior treatment decisions and to anticipate the likely downstream order trajectory, thereby supporting order-aware care planning and discharge readiness assessment.

\textbf{Department consultation synthesis.}
These tools retrieve specialist consultation records from HIS. Their role is to expose cross-disciplinary expert opinions and to support complex cases with multimorbidity or perioperative concerns. The consultation environment spans 609 specialty categories, organized into:
\begin{itemize}
    \item preoperative assessment consultations, including anaesthesiology evaluation and preoperative medical risk stratification;
    \item perioperative management consultations, including critical care and complication management;
    \item rehabilitation planning, including functional recovery planning and supportive care;
    \item specialty-specific consultations, including cardiology, neurology, endocrinology, respiratory medicine, infectious disease, dermatology and urology.
\end{itemize}
The outputs are consultation notes and specialist recommendations that can be integrated into subsequent planning, discharge and rehabilitation decisions.

\textbf{Similar case retrieval.}
This module retrieves analogous historical musculoskeletal care cases from the institutional similar-case retrieval resource. This resource contains 48{,}464 earlier in-house cases from 2004 to 2011, reserved by calendar time and kept separate from benchmark curation, as described in Appendix~\ref{sec:retrospective_dataset_curation}. Its function is to identify prior cases sharing similar phenotypic presentations, imaging patterns, or treatment trajectories, thereby providing empirical reference in uncertain or atypical situations. The retrieval operates through two complementary strategies:
\begin{itemize}
    \item {Diagnosis-based retrieval}, which matches cases by ICD-10 code and diagnosis name similarity, ensuring coverage of cases with concordant nosological labels;
    \item {Text-based retrieval}, which computes term frequency--inverse document frequency (TF-IDF) similarity over clinical narratives to identify phenotypically analogous cases beyond strict diagnostic label matching.
\end{itemize}
Given a patient-specific query, the module returns the top-$k$ most similar historical cases together with their trajectory summaries, which serve as empirical evidence for diagnosis, treatment selection, and prognosis estimation.

\paragraph{External knowledge world.}
The external knowledge world serves as the evidence-based complement to patient-specific hospital evidence. Its role is to provide contemporary, standardized, and globally shared medical knowledge that cannot be guaranteed to reside within the model parameters alone. The inputs to this world are task-specific clinical queries and intermediate reasoning hypotheses; the outputs are external evidence fragments, structured relations, or reference passages that support evidence-grounded synthesis.

The aggregated external evidence is denoted by:
\begin{equation}
\mathcal{K}_{\mathrm{ext}}(q) = \bigcup_{m=1}^{M} \mathcal{K}_{m}(q),
\end{equation}
where $\mathcal{K}_{m}(\cdot)$ denotes the $m$-th external retrieval module queried by clinical query $q$, and $M$ is the number of external evidence sources. In practice, this world comprises the following retrieval modules.

\textbf{Diagnostic guideline retrieval.}
This module retrieves guideline-level evidence for diagnosis and management questions, anchoring OrthoPilot's recommendations to authoritative clinical consensus. The guideline corpus contains 1{,}604 guideline documents and is accessed through two specialized search tools:
\begin{itemize}
    \item {Guideline search}, which dynamically queries a prioritized set of authoritative sources, including society recommendations, expert consensus statements and guideline-associated reference documents, and terminates once sufficiently relevant evidence has been identified, thereby reducing redundancy while preserving timeliness and clinical specificity;
    \item {Differential diagnosis search}, which accepts a structured symptom list with optional patient age and specialty constraints, and returns systematically organized differential-diagnostic evidence to support exclusion-based reasoning.
\end{itemize}
The returned evidence consists of relevant recommendation statements, evidence summaries and section-level passages from guideline documents, rather than abstracts alone. These materials inform downstream reasoning on disease definition, subtype classification, operative indication and treatment selection. In this way, the module serves as the primary interface through which OrthoPilot aligns its decisions with standardized clinical pathways.

\textbf{Real-time web search and academic retrieval.}
This module provides real-time global medical context through a combination of general web search, encyclopaedic retrieval, and academic literature search. Its role is to complement guideline- and literature-based evidence by supplying broader background knowledge, especially in situations where the agent must rapidly interpret symptoms, clarify terminology, or expand a differential diagnosis. Given a clinical query, the module operates through three complementary channels:
\begin{itemize}
    \item {Web search} (via Serper/Google API), which emphasizes timeliness and breadth, retrieving recently published medical information, differential-diagnostic clues, and emerging treatment paradigms. Retrieved web pages are further processed through a goal-conditioned extraction pipeline powered by a lightweight language model, which summarizes and filters content relevant to the original clinical query;
    \item {Wikipedia retrieval}, which emphasizes concept-level grounding, providing structured explanatory content on anatomical definitions, pathophysiological mechanisms, and general interpretive context;
    \item {Semantic Scholar search}, which provides access to the academic literature graph containing over 200 million publications, returning citation-linked metadata (title, authors, year, venue, abstract) for research articles relevant to the query, thereby complementing web-scale search with scholarly depth and bibliometric context.
\end{itemize}
Operationally, this module functions as a lightweight global knowledge interface: web search contributes recency and coverage, Wikipedia contributes conceptual organization, and Semantic Scholar contributes citation-aware academic retrieval. Importantly, the module does not rely solely on the summary-level information returned by search engines or academic search APIs. When more detailed evidence is needed, it can further follow retrieved links and fetch the underlying webpages or documents for deep browsing, enabling extraction of source-specific passages, contextual details, guideline statements, and fine-grained evidence beyond the initial search summaries. In this way, search serves as a discovery layer, while deep webpage fetching supports evidence verification and more detailed clinical grounding.

\textbf{Multi-source biomedical literature retrieval.}
This module provides systematic, research-oriented retrieval augmentation for higher-level evidence needs. Built on the MedRAG \methodcite{xiong2024benchmarking}{50} multi-corpus retrieval framework, it supports parallel evidence aggregation across five complementary biomedical corpora:
\begin{itemize}
    \item {PubMed}, which provides access to peer-reviewed biomedical abstracts, supporting evidence retrieval for epidemiology, clinical outcomes, therapeutic efficacy and mechanistic studies;
    \item {StatPearls}, which provides clinically structured, peer-reviewed summaries authored by domain experts, bridging primary research evidence and day-to-day clinical practice with emphasis on diagnostic criteria, treatment algorithms and management protocols;
    \item {Medical textbooks}, which provide curated textbook passages covering foundational disease knowledge, classification systems, diagnostic criteria and treatment principles across core clinical specialties;
    \item {Wikipedia}, which serves as an auxiliary corpus for rapid contextualization of terminology and biomedical concepts when domain-specific sources lack coverage;
    \item {MedCorp}, which provides additional biomedical passages for retrieval across heterogeneous clinical and scientific topics.
\end{itemize}
For each query $q$, the module independently retrieves the top-$k$ relevant passages from each corpus and returns a merged, deduplicated evidence set. The retrieval combines sparse lexical matching with BM25 \methodcite{Robertson2009ThePR}{51} and dense semantic retrieval to improve coverage across heterogeneous evidence types. Formally, the aggregated evidence is:
\begin{equation}
\mathcal{E}_{\mathrm{lit}}(q) = \bigcup_{c \in \mathcal{C}} \mathrm{Top\text{-}k}\bigl(\mathrm{BM25}(q, \mathcal{D}_c) \cup \mathrm{Dense}(q, \mathcal{D}_c)\bigr),
\end{equation}
where $\mathcal{C} = \{\text{PubMed}, \text{StatPearls}, \text{Textbooks}, \text{Wikipedia}, \text{MedCorp}\}$ denotes the corpus set and $\mathcal{D}_c$ denotes the document collection of corpus $c$. By aggregating ranked passages across multiple corpora, this module returns evidence suitable for retrieval-augmented reasoning in evidence-sensitive clinical tasks.

\textbf{Hierarchical textbook semantic retrieval.}
This module retrieves stable, structured medical knowledge from a curated corpus of open medical textbooks covering 14 core clinical specialties. Its role is to provide high-reliability background knowledge on orthopaedics, imaging, complications, rehabilitation and related topics. These evidence types require conceptual completeness and structural coherence beyond what snippet-level retrieval can provide. The textbook corpus is organized hierarchically by specialty, chapter and subsection. Given a clinical query $q$, the system performs coarse-to-fine semantic matching over this three-level hierarchy:
\begin{itemize}
    \item {Specialty selection}: the query embedding is compared against specialty-level representations to identify the most relevant clinical domains;
    \item {Chapter localization}: within the selected specialty, chapters are ranked by semantic similarity to the query;
    \item {Passage retrieval}: the most relevant subsections within the top-ranked chapters are extracted and returned as contiguous evidence passages.
\end{itemize}
This hierarchical design reduces the effective search space from the full multi-specialty corpus to a targeted chapter-level neighbourhood, preserving the structural coherence and didactic organization of textbook knowledge that flat keyword-based or chunk-level retrieval would disrupt.

\textbf{Clinical decision support libraries.}
This module retrieves case-based decision-support evidence from PMC-Patients and RareArena. It is not tied to a single care stage, but instead provides external patient-level evidence that can support diagnostic, perioperative, therapeutic and follow-up decisions. In implementation, the service maintains FAISS indices for ReCDS-PPR, ReCDS-PAR, RareArena-RDS and RareArena-RDC, with queries encoded by sentence-transformer embeddings against task-specific corpora. The tools provide three complementary retrieval modes:
\begin{itemize}
    \item {Patient-to-patient retrieval} (PPR), which searches the ReCDS-PPR reference corpus containing 155.2k patient summaries from PMC-Patients and returns cases with similar clinical narratives;
    \item {Patient-to-article retrieval} (PAR), which matches the input patient summary against the ReCDS-PAR PubMed article corpus, retrieving article-level evidence associated with phenotypically similar presentations;
    \item {Rare disease case retrieval} (RareArena), which searches rare disease case corpora derived from PMC-Patients. It supports two information-truncation settings: {RDS} (rare disease screening), in which cases are truncated before disease-defining diagnostic tests for early differential reasoning, and {RDC} (rare disease confirmation), in which cases are truncated up to the final diagnosis for confirmation-stage reasoning. Cases with potential diagnosis leakage are removed during corpus construction.
\end{itemize}
RareArena is particularly important for expanding OrthoPilot's diagnostic coverage beyond common musculoskeletal conditions. Atypical bone tumours, metabolic bone disease, hereditary skeletal dysplasia and rare connective-tissue disorders may be sparse in institutional cohorts or mainstream evidence sources. By adding patient-level evidence from a dedicated rare disease collection, this module supports reasoning about uncommon differential diagnoses and reduces anchoring on high-prevalence conditions.

\textbf{Structured medical knowledge graph retrieval.}
This module complements unstructured document retrieval with relation-aware retrieval over CPubMed, a large-scale typed medical knowledge graph. The graph contains 4,580,006 triples and 1,810,772 unique entities, covering 45 standardized relation types and 12 semantic entity categories. These relation types span therapeutic, diagnostic, symptomatic, epidemiological and prognostic categories. The module supports explicit retrieval of standardized associations among diseases, drugs, symptoms, examinations, anatomical sites, prognostic factors and treatment procedures. In implementation, the deployed tool set includes two core query tools, three utility tools and 45 relation-specific wrappers. For manuscript clarity, these interfaces can be summarized into three functional modes:
\begin{itemize}
    \item {Entity-relation query}, which retrieves triples associated with a given medical entity, optionally filtered by a specified relation type or entity type;
    \item {Entity relation profiling}, which returns the relation-type distribution and statistics for a given entity, enabling the agent to inspect the entity's relational neighbourhood before issuing targeted queries;
    \item {Fuzzy entity matching}, which maps informal, abbreviated or misspelled entity names to standardized knowledge graph entries with ranked similarity scores, supporting robust query formulation from noisy clinical inputs.
\end{itemize}
Formally, given an entity-centric query $q$, an optional relation constraint $r_f$ and an optional type constraint $\alpha_f$, the returned structured evidence set is:
\begin{equation}
\begin{aligned}
\mathcal{S}(q, r_f, \alpha_f) = \bigl\{\tau_i \in \mathcal{T}\ \bigm|\;&
q \in \{h_i, u_i\},\\[-0.15em]
&(r_f = \varnothing \lor r_i = r_f),\\[-0.15em]
&(\alpha_f = \varnothing \lor \alpha_i(q) = \alpha_f)\bigr\},
\end{aligned}
\end{equation}
where $\tau_i = (h_i, r_i, u_i, \alpha_i^{(h)}, \alpha_i^{(u)})$ denotes a typed triple, $h_i$ and $u_i$ denote the head and tail entities, $r_i$ denotes the relation type, and $\alpha_i(q)$ denotes the entity-type label of the matched query entity. This design lets the agent navigate the knowledge graph even when the input entity name is ambiguous or non-standard. The agent can first resolve the entity through fuzzy matching, then profile its relational structure and finally issue targeted relation queries to extract the medical facts needed for downstream reasoning.

In practice, the in-hospital information world and the external knowledge world are not used independently, but are jointly orchestrated through Tool Plaza to support evidence-grounded reasoning across the full musculoskeletal care pathway.

During retrospective evaluation, OrthoPilot used the same Tool Plaza architecture as in deployment, while patient-specific evidence returned by in-hospital tools was constrained to the information available at the evaluated clinical time. Imaging reports, laboratory results, pathology reports, order records and consultation notes can occur before or after surgery according to the actual care trajectory. For benchmark evaluation, each record was manually aligned to the reconstructed timeline and exposed only when its timestamp preceded the evaluated node. In clinical deployment, the same principle is implemented through real-time HIS updates, so newly generated evidence becomes available as soon as it enters the authorized clinical record.

For a patient $p$ evaluated at pathway node $n$ with decision time $t_n$, the evidence returned by in-hospital tool $j$ was filtered as:
\begin{equation}
\mathcal{E}_{j,n}(p)
=
\left\{
 x \in \mathcal{D}_j(p)
 \;\middle|\;
 \mathrm{time}(x) \leq t_n,\;
 x \notin \mathcal{M}_n
\right\},
\end{equation}
where $\mathcal{D}_j(p)$ denotes the patient-specific evidence store queried by tool $j$, $\mathrm{time}(x)$ is the clinical timestamp of evidence item $x$, and $\mathcal{M}_n$ denotes node-specific masked content that could reveal the target answer. Masked content includes diagnosis labels corresponding to the evaluated diagnostic node and task outputs that would be generated only after that decision point.

This filtering rule prevented answer leakage during OrthoPilot evaluation while preserving the temporal evidence envelope of real clinical care. The same in-hospital tool could therefore be invoked at multiple nodes, but returned only the subset of patient evidence generated before $t_n$. Imaging, laboratory, pathology, order and consultation tools could contain evidence relevant to both preoperative and postoperative reasoning, with visibility determined by record timestamp and node-specific masking. For the MDT task, consultation notes containing the target specialist recommendation were withheld. Earlier non-target consultation records were retrieved only if their timestamps preceded the evaluated node. After an MDT event had occurred, its timestamped consultation note could support later tasks. The historical similar-case retrieval resource was available across nodes, but the query used to retrieve analogous cases was constructed only from time-valid patient evidence available at $t_n$. External knowledge tools did not contain patient-future information and remained available across tasks, but their outputs were integrated only with the time-valid in-hospital evidence available at that node.

\subsubsection{Agent reasoning framework}

Longitudinal clinical management requires each recommendation to be made within the evidence available at that point in the care pathway, while remaining coherent with prior decisions and downstream clinical goals. OrthoPilot formalizes this process as a ReAct-style reasoning loop over Tool Plaza. For each request, OrthoPilot receives the query $q$, the current patient state $s$, and the corresponding pathway node. It then alternates between deliberation, action selection, and evidence integration until the available evidence is sufficient for a structured recommendation or the predefined turn budget is reached.

To standardize access to heterogeneous clinical resources, we define {Tool Plaza} as a unified abstraction layer over all callable tools:
\begin{equation}
\mathcal{T} = \mathcal{T}_{\mathrm{in}} \cup \mathcal{T}_{\mathrm{ext}} \cup \mathcal{T}_{\mathrm{mm}},
\end{equation}
where $\mathcal{T}_{\mathrm{in}}$ denotes in-hospital tools, $\mathcal{T}_{\mathrm{ext}}$ denotes external knowledge tools, and $\mathcal{T}_{\mathrm{mm}}$ denotes multimodal tools. Tool Plaza exposes clinical databases, similar-case retrieval, knowledge graph search, medical retrieval, web search, and multimodal parsing through a common message-passing interface. This allows OrthoPilot to invoke patient-specific information and external evidence through the same executable framework.

At the start of a query, OrthoPilot initializes the in-hospital evidence pool $\mathcal{X}_{\mathrm{in}}$, the external evidence pool $\mathcal{K}_{\mathrm{ext}}$, and the trajectory history $H_0$. The multi-turn trajectory is represented as:
\begin{equation}
H_T = (\tau_0, a_0, o_0, \tau_1, a_1, o_1, \ldots, \tau_T, a_T),
\end{equation}
where $\tau_t$ denotes the reasoning trace at step $t$, $a_t$ denotes the selected action, and $o_t$ denotes the observation returned after action execution. At each step, OrthoPilot reasons over $q$, $s$, $\mathcal{X}_{\mathrm{in}}$, $\mathcal{K}_{\mathrm{ext}}$ and $H_t$, then selects one action from $\mathcal{T}$. In-hospital actions update $\mathcal{X}_{\mathrm{in}}$, external knowledge actions update $\mathcal{K}_{\mathrm{ext}}$, and multimodal actions convert images or documents into structured patient evidence.

We impose a sequential tool-use constraint so that only one action is executed at each reasoning step. This constraint makes the reasoning trace auditable and preserves the temporal evidence boundary of each pathway node. The same framework is used during benchmark evaluation and clinical deployment. In evaluation, it prevents future information from entering early-stage reasoning. In deployment, it maintains a transparent record of how patient facts and external evidence are accumulated before final synthesis.

\subsubsection{Application workflow and deployment}

We developed a clinical web application to translate OrthoPilot from an executable reasoning framework into longitudinal clinical management support. The application organized each interaction as a persistent case session rather than a single isolated query. This allowed clinicians to initiate a task, revisit prior evidence, continue discussion across care stages, and preserve the clinical context needed for management decisions over time.

\begin{figure}[h]
\centering
{%
\includegraphics[width=\textwidth,height=\textheight,keepaspectratio]{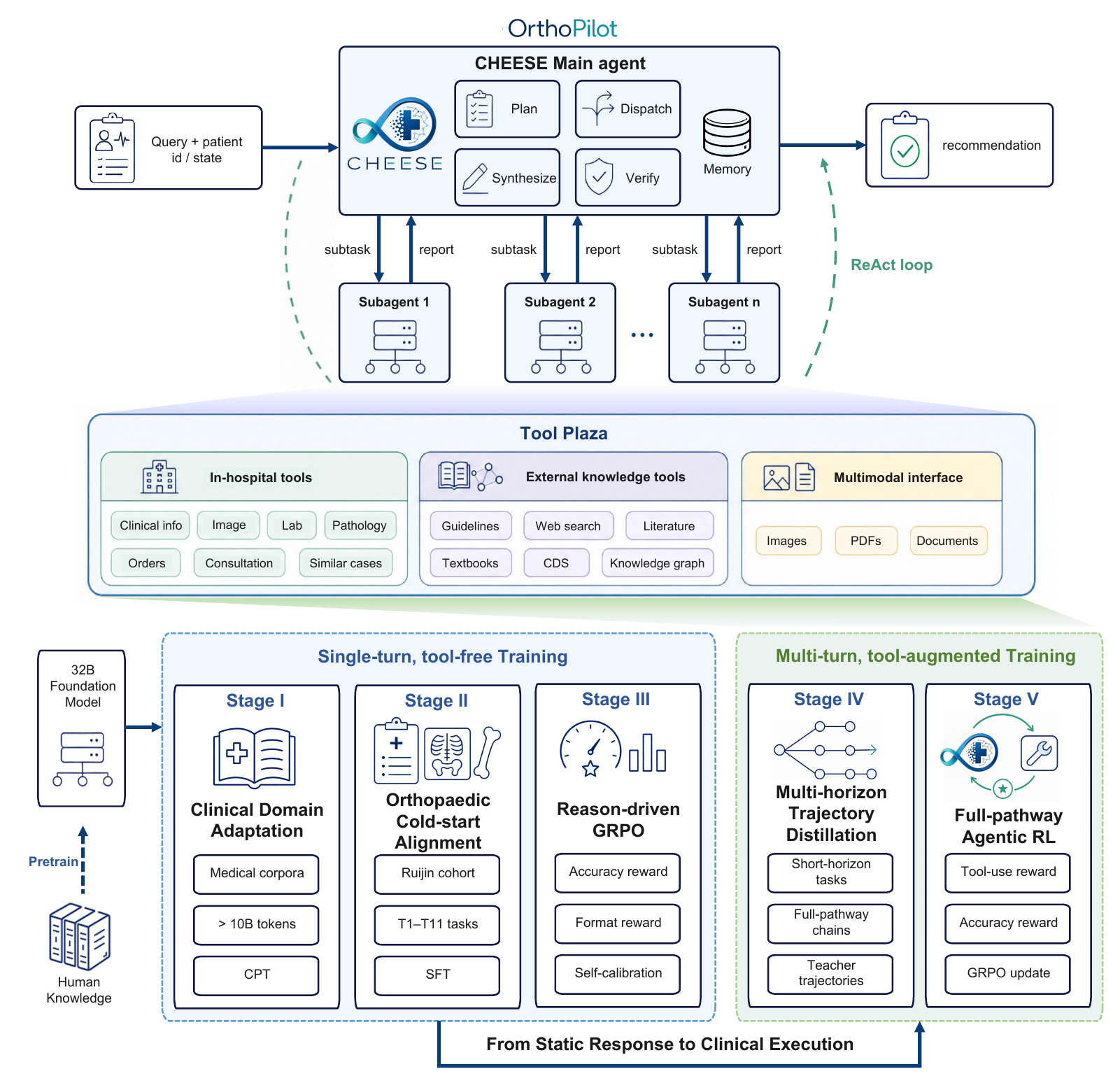}  %
}
\caption{\textbf{Hierarchical agent workflow and training curriculum of OrthoPilot.}
\textbf{a,} CHEESE functions as the main clinical reasoning agent. It receives the patient query and current state, identifies missing evidence and selects actions from Tool Plaza. Subagents operate in isolated sessions to retrieve in-hospital patient data, external knowledge and multimodal reports. These structured reports are returned to CHEESE for iterative synthesis and final recommendation generation.
\textbf{b,} The CHEESE training curriculum progressively transforms a 32B foundation model into a clinical reasoning agent. Stages I--III involve single-turn, tool-free training for domain adaptation, musculoskeletal cold-start alignment and reason-driven GRPO optimization. Stages IV--V implement multi-turn, tool-augmented training for multi-horizon trajectory distillation and full-pathway agentic reinforcement learning. The curriculum incorporates short- and long-horizon supervision, task- and pathway-level trajectories, and rewards for accuracy, format compliance and effective tool usage.}
\label{fig:orthopilot_agent_workflow}
\end{figure}

In deployment, CHEESE served as the main agent that maintained the clinical objective, pathway position, patient state, and accumulated trajectory $H_t$ (Fig.~\ref{fig:orthopilot_agent_workflow}). Actions selected by the ReAct loop could be executed directly through Tool Plaza or delegated to a subagent when focused evidence acquisition was required. Each subagent received a scoped subtask $q_t$ and the tools enabled for that session. It then retrieved in-hospital evidence, external knowledge, or multimodal reports, and returned a structured report rather than a raw tool transcript. CHEESE treated this report as the observation $o_t$, updated the evidence pools, and retained control of stopping and final synthesis. The single-query procedure is formalized in Algorithm~\ref{alg:orthopilot_workflow} in the Appendix.

Users could initiate tasks either from predefined pathway entries or from free-text queries. For hospital deployment, a request could be anchored to a patient identifier, which allowed OrthoPilot to assemble patient-specific evidence from the in-hospital information world, including longitudinal electronic health records, laboratory results, imaging reports, operative records and other clinical documentation. The interface also accepted supplementary multimodal materials, including typed notes, images, PDF documents and text files. These inputs were parsed on the server and incorporated into the query context. Access control followed role-specific permissions, so users could access only the information and tools authorized for their role and scope.

Before execution, users configured session-specific capabilities through a marketplace interface that displayed available tools and skills, their definitions and tool dependencies. The interface supported adding, removing, enabling or disabling tools and custom skills. In agent mode, CHEESE invoked the selected capabilities, dispatched subagents when needed, and streamed concise reasoning summaries, tool calls and execution progress to the frontend in real time. Retrieved evidence from the in-hospital information world, the external knowledge world and multimodal inputs was synthesized into a structured recommendation for clinical review, patient communication and downstream documentation.

\FloatBarrier

\FloatBarrier

\begin{figure}[h]
\centering
{%
\includegraphics[width=\textwidth,height=\textheight,keepaspectratio]{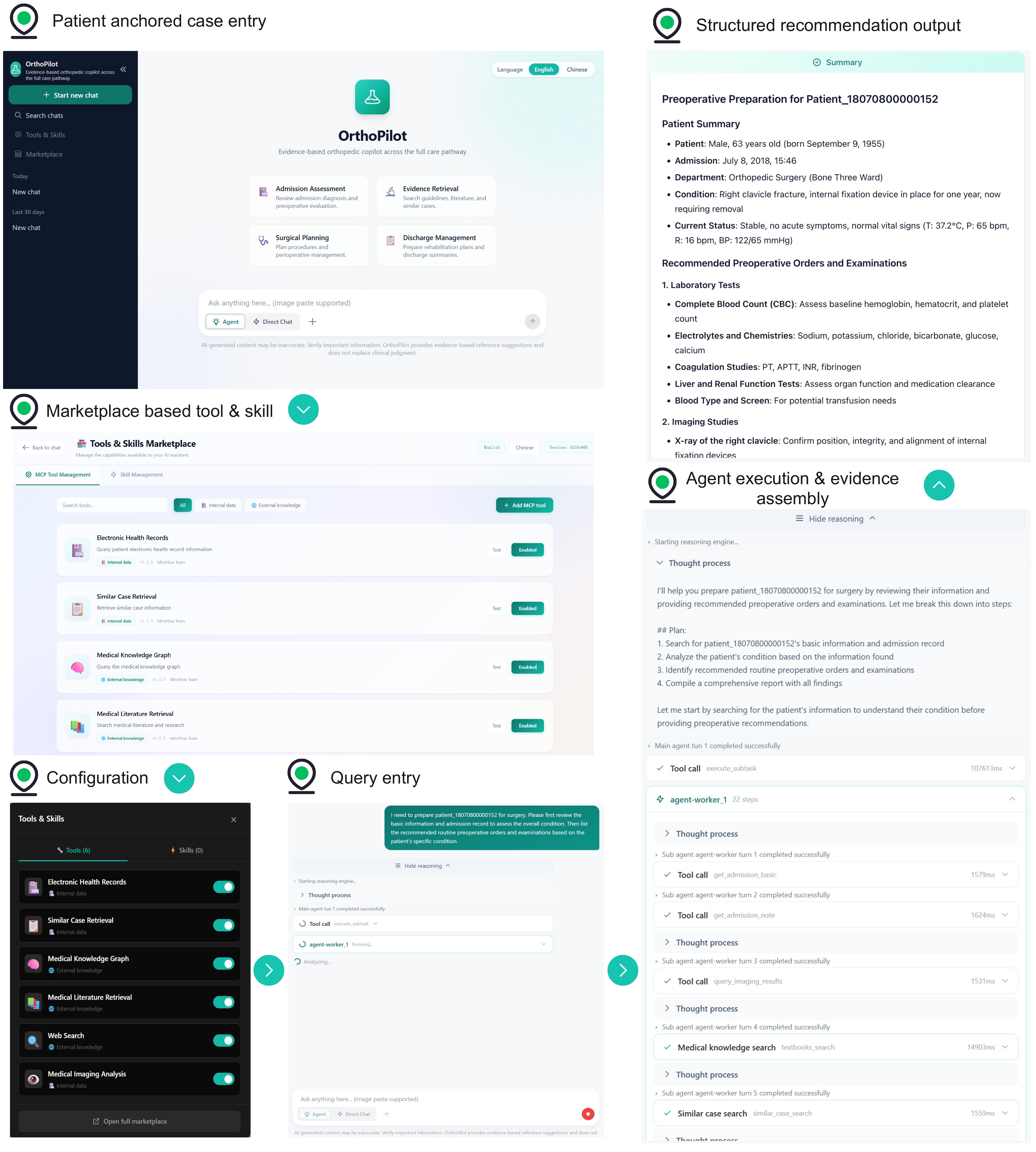}%
}
\caption{\textbf{Frontend workflow of OrthoPilot across the musculoskeletal care pathway.}
The figure illustrates a representative application workflow of OrthoPilot. A clinician initiates a patient-anchored task through case entry. Session-specific tools and skills are configured through the marketplace and sidebar selector. OrthoPilot then streams intermediate reasoning, tool calls, subagent activity and evidence retrieval from hospital records and external knowledge resources. The assembled evidence is synthesized into a structured recommendation for clinical review.}

\label{fig:frontend_workflow}
\end{figure}

\FloatBarrier

\subsection{Study design}

\subsubsection{Retrospective clinical study}

We collected de-identified retrospective data from the Ruijin Hospital Orthopaedic Cohort as the internal longitudinal cohort. This cohort included 183{,}472 consecutive patients receiving musculoskeletal care at Ruijin Hospital, Shanghai, China, between 2004 and 2024. Available data included electronic health records, imaging studies and reports, laboratory results, pathology reports, consultation records, operative documentation, medical orders, discharge summaries and follow-up information. Retrospective data use followed institutional ethics approval, with waiver of informed consent for de-identified records. Detailed inclusion and exclusion criteria, cohort allocation, metadata definitions and task construction are provided in Appendix Section~\ref{sec:details_datasets}.

To evaluate OrthoPilot on retrospective data, we conducted four prespecified analyses. First, we constructed a fixed internal evaluation cohort from the Ruijin longitudinal cohort using patient-level separation from model-development data. This cohort was used to instantiate the 11 full-pathway tasks described in Methods Section~\ref{sec:evaluation_framework_benchmark}. Second, we assembled an independent external multicentre validation cohort from 60 participating hospitals to test generalizability across institutions. External data were reserved for validation only, and no model parameters were tuned or adjusted using these data.

Third, we performed a disease-level retrospective analysis within the internal evaluation cohort. Disease codes were stratified by distributional status and prevalence to assess common, rare, seen and previously unseen disease groups under the same evaluation framework. Fourth, we conducted a retrospective human-machine reader study using Ruijin surgical musculoskeletal cases that were excluded from model training and internal evaluation. Physicians from three experience tiers reviewed the same cases without assistance and then repeated the assessments after a washout interval with OrthoPilot output available as a reference. OrthoPilot was also evaluated as a standalone reader. The detailed reader-study protocol and statistical analysis are described in Appendix Section~\ref{sec:reader_study}.

\subsubsection{Prospective clinical study}
To prospectively evaluate the clinical effectiveness of OrthoPilot, we conducted two independent prospective studies at Ruijin Hospital, Shanghai, China, including a crossover reader study focusing on physician decision-making and a randomized clinical deployment study evaluating system-wide clinical impact. Written informed consent was obtained from participating physicians and from patients who contributed prospectively collected clinical information or patient-reported outcomes. Patient data were de-identified prior to analysis. The prospective Ruijin cohorts are entirely independent of the retrospective cohort used for model development, with no overlap in patients or clinical records. All analyses followed prespecified evaluation criteria defined before study initiation.

\paragraph{Prospective crossover reader study}
To evaluate the impact of OrthoPilot on clinical decision-making across key musculoskeletal tasks, we conducted a prospective crossover reader study using 1,870 complex musculoskeletal cases from Ruijin Hospital. Consecutive cases between September 2025 and April 2026 were screened from the institutional clinical data warehouse. Eligible cases were defined as patients requiring multi-step musculoskeletal decision-making involving diagnostic evaluation, surgical planning or intraoperative reassessment. Each case included multimodal clinical data, including radiographic imaging (X-ray, CT or MRI), physical examination findings, laboratory tests, operative records when available and longitudinal clinical notes. Cases with incomplete imaging data, missing reference diagnoses or insufficient clinical documentation were excluded.
Adjudication criteria for each clinical task were predefined before study initiation and are summarized in Supplementary Table \ref{tab:prospective_cohort_reader_study}.

A total of 30 orthopaedic physicians were prospectively recruited and stratified into three experience-level groups of ten: junior ($<$5 years), mid-level (5--10 years), and senior ($>$10 years). Each physician independently completed four predefined clinical decision tasks for every assigned case: admission diagnosis (P1), preoperative diagnosis (P2), surgical planning (P3) and intraoperative diagnosis (P4).

Within each group, the 1,870 cases were distributed equally among the ten physicians, with each physician assigned approximately 187 cases. Each physician's assigned cases were then randomly divided into two balanced subsets (Set~A and Set~B), stratified by disease category and complexity. In the first phase, each physician evaluated Set~A with OrthoPilot assistance and Set~B without assistance, completing all four tasks for each case. After a washout period of at least six weeks to minimise recall bias, each physician reassessed the same assigned cases under the reversed condition, evaluating Set~A without assistance and Set~B with OrthoPilot assistance. This within-physician crossover generated paired observations under both conditions for every physician-case combination. All three experience-level groups followed the same protocol independently, enabling prespecified subgroup analyses stratified by physician experience.
Diagnostic performance for P1, P2 and P4 was evaluated using accuracy against reference standards. Surgical planning quality for P3 was assessed using a blinded expert committee composed of fifteen senior orthopaedic surgeons with more than 15 years of clinical experience, with each plan independently evaluated by at least three of them across five predefined dimensions (Supplementary Table \ref{tab:prospective_cohort_reader_study}), including surgical approach selection, fixation strategy, risk preparedness, patient-specific adaptation and plan feasibility. Each dimension was rated on a 10-point Likert scale, and the final score was calculated as a weighted composite according to prespecified criteria (Supplementary Tables \ref{tab:surgical_planning_criteria} and \ref{tab:score_interpretation}). For each case under each condition, correctness for P1, P2 and P4 was determined by majority vote across the three experience-tier physicians who evaluated that case, and P3 was represented by its case-level quality category. Full-chain outcomes were therefore defined at the case level (denominator = 1,870 cases): full-chain success required P1, P2 and P4 to be all correct and P3 to be classified as high quality, whereas failure-prone trajectories were defined as cases with at least one incorrect decision among P1, P2 and P4.

\paragraph{Prospective randomized clinical deployment study}

To evaluate the real-world clinical impact of OrthoPilot, we conducted a single-centre, prospective randomized controlled study at Ruijin Hospital over an 8-month period from September 2025 to April 2026. A total of 8,240 consecutive musculoskeletal inpatients admitted for diagnostic assessment or surgical treatment were enrolled. Adult patients undergoing musculoskeletal evaluation requiring diagnostic assessment, surgical planning, postoperative management or rehabilitation planning were eligible. Exclusion criteria included emergency trauma requiring immediate surgery, incomplete electronic medical records and hospital stay shorter than 24 hours.
Patients were randomly assigned in a 1:1 ratio to either the standard-care group or the OrthoPilot-assisted group using computer-generated randomization stratified by major disease category, surgical complexity and admission type. In the control group, patients were managed according to the conventional clinical workflow without AI assistance. In the intervention group, OrthoPilot was integrated into the hospital information system and provided structured decision support before final clinical decision-making, while all final decisions remained under physician responsibility. OrthoPilot operated on multimodal clinical data, including electronic medical records, radiographic imaging, laboratory results and real-time vital signs, and supported decision-making across the full musculoskeletal care pathway, including pre-diagnostic evaluation, surgical decision and execution, postoperative management and rehabilitation planning.

Prospective outcomes were evaluated across four predefined domains, including hospital-level efficiency, physician workload, nursing workflow and patient-reported experience. Hospital-level metrics, including length of stay and bed turnover rate, were extracted from hospital administrative databases. Bed turnover was calculated over four consecutive 2-month intervals during the study period to account for temporal variation in patient flow.
Physician workload was assessed using the validated NASA Task Load Index (NASA-TLX), administered at predefined timepoints during clinical workflow (Supplementary Table \ref{tab:nasa_tlx}). A total of 42 physicians participating in the study completed the workload assessment under both study conditions. Documentation efficiency was prospectively recorded using structured time-logging forms embedded within the electronic medical record system, with all participating physicians contributing data across admission notes, progress notes and discharge summaries (Supplementary Table \ref{tab:physician_documentation}).
Nursing workflow outcomes, including documentation burden and response time, were collected using standardized nurse questionnaires and time-tracking logs. A total of 75 registered nurses involved in inpatient care during the study period completed the questionnaires (Supplementary Table \ref{tab:nursing_workflow}). Patient-reported outcomes, including satisfaction, communication quality and accessibility of health information, were assessed using structured Likert-scale questionnaires administered at discharge. All eligible patients were invited to participate, and responses were obtained from 7,824 of 8,240 enrolled patients (response rate 94.95\%) (Supplementary Tables \ref{tab:patient_satisfaction}, \ref{tab:communication_quality}, \ref{tab:health_information}).
Outcome definitions and statistical analyses were prespecified before study initiation. Investigators responsible for outcome assessment were blinded to group allocation whenever feasible, and data completeness and protocol adherence were monitored throughout the study period.

\subsection{Evaluation Framework and Benchmark}
\label{sec:evaluation_framework_benchmark}

In order to rigorously evaluate the clinical reasoning capabilities of OrthoPilot, we developed a comprehensive evaluation framework that includes a meticulously constructed benchmark dataset. This benchmark, OrthoBench, is designed to assess both individual task-level performance and the model's ability to maintain decision consistency across the entire musculoskeletal care pathway. The dataset encompasses diverse clinical data from real-world sources and enables evaluation in a manner that mirrors the complexities encountered in actual clinical workflows.

\subsubsection{Benchmark Construction and Dataset Sources}

OrthoBench was derived from the Ruijin Hospital longitudinal cohort of patients receiving musculoskeletal care between 2004 and 2024. This source cohort included 183{,}472 patients and more than 100 million data points. Earlier in-house cases from 2004 to 2011 were reserved for the historical similar-case retrieval resource and kept separate from benchmark curation. The remaining eligible cohort was allocated at the patient level, yielding a fixed OrthoBench test cohort of 5{,}905 patients across 1,000 disease categories. Detailed cohort curation, exclusion criteria and patient-level allocation are provided in Appendix Section~\ref{sec:details_datasets}.

Although the evaluation targeted full-pathway clinical capability, OrthoBench was organized as node-specific question-answering tasks. Each QA instance corresponded to one of the 11 predefined care-pathway nodes and was constructed from a reconstructed patient timeline. This design allowed all systems to be evaluated at the same clinical decision point with the same information boundary, preventing unfair comparisons caused by unequal access to future records or target answers.

During OrthoPilot evaluation, patient-specific information was not supplied as a single static prompt. Instead, the current pathway node defined the decision time, and OrthoPilot retrieved time-valid in-hospital evidence through Tool Plaza under the masking rules described above. For non-agentic models, we constructed static prompts from the same time-valid evidence envelope. Thus, differences in performance reflected model reasoning and evidence-use capability rather than differences in available patient information. For example, preoperative-diagnosis prompts contained the admission record and available preoperative imaging, laboratory results, orders and other records generated before that decision time. They did not contain operative findings, postoperative records, discharge diagnoses or other future information. The external multicentre cohort from 60 hospitals, comprising 6{,}796 cases, was processed with the same task definitions and information-boundary rules, and was used exclusively as an independent external validation set to assess generalizability across institutions, patient populations and clinical practice patterns.

\paragraph{Benchmark Construction Process}

OrthoBench was constructed by assigning each eligible patient to the 11 evaluation nodes defined in Table~\ref{tab:task_node_mapping}. Tasks~1--4 correspond to diagnostic reasoning, comprising admission diagnosis, preoperative diagnosis, intraoperative diagnosis and discharge diagnosis. Tasks~5--11 correspond to clinical management, comprising perioperative assessment, surgical planning, preoperative orders, postoperative orders, discharge summary, rehabilitation planning and multidisciplinary consultation. For each node, the input was assembled from the reconstructed patient timeline under the stage-specific information constraints described above, and the output was defined as the corresponding diagnosis, order set, assessment, operative plan, summary, rehabilitation plan or consultation synthesis. This construction preserved the full-pathway structure while allowing each node to be evaluated as a standardized QA instance.

The resulting OrthoBench test set contains 135{,}745 evaluation instances from 5{,}905 patients, covering 11 clinical tasks and 1{,}000 disease categories. Diagnostic tasks include open-ended, multiple-choice and true-or-false formats, whereas clinical management tasks are evaluated as open-ended generation. Disease categories are stratified by distributional status and prevalence to support analyses of in-distribution, out-of-distribution, rare and non-rare conditions. Detailed task counts, disease taxonomy, generalization strata and evaluation metrics are provided in Appendix Section~\ref{sec:orthobench_design_metrics}.

\subsubsection{ORACLE: automated evaluation for open-ended clinical reasoning}

Evaluating open-ended clinical question answering is fundamentally different from evaluating conventional short-form generation tasks. Standard text-overlap metrics, such as BLEU and ROUGE, primarily measure surface n-gram similarity between a model response and a reference answer, but they are poorly aligned with clinical evaluation. Response quality depends on whether the answer captures the essential clinical facts, diagnostic rationale, management priorities and actionable recommendations required for patient care.

To address this limitation, we developed ORACLE (Open Response Assessment for Clinical Language Evaluation), a structured evaluation framework for open-ended clinical tasks. ORACLE uses physician-defined rubrics to convert reference answers into scorecards of clinically salient entities and content items. It then evaluates whether these rubric-defined entities and items are covered in the model response, providing a structured alternative to surface-form similarity. A case-level example of this evaluation workflow is shown in Fig.~\ref{fig:oracle_validation}b, and the scoring procedure is summarized in Algorithm~\ref{alg:oracle_scoring}.

For each sample, task-specific criteria and the reference answer were converted into a structured scorecard. The scorecard was organized by clinical category and by importance level, with levels $\mathcal{L}=\{\mathrm{primary},\mathrm{secondary},\mathrm{additional}\}$. The rubric-defined entity set was defined as:
\begin{equation}
\mathcal{E} = \{e_{ilj} \mid i=1,\ldots,m;\; l\in\mathcal{L};\; j=1,\ldots,n_{il}\},
\end{equation}
where $e_{ilj}$ denotes the $j$th key entity or content item in category $i$ and importance level $l$. Empty criterion fields were skipped during scoring.

Given a model-generated answer $A$, LLM judges assessed whether each rubric-defined entity or item was covered by the response. The resulting binary coverage decision was denoted as $c(e_{ilj}, A)\in\{0,1\}$. When multiple judges were used, item-level coverage decisions were merged by majority vote before scoring. The normalized sample score was computed as:
\begin{equation}
\mathrm{Score}(A) =
\frac{
\sum_{i=1}^{m}\sum_{l\in\mathcal{L}}\sum_{j=1}^{n_{il}}
w_l \, c(e_{ilj}, A)
}{
\sum_{i=1}^{m}\sum_{l\in\mathcal{L}}
n_{il} w_l
},
\end{equation}
where $w_l$ denotes the task-specific grading weight for importance level $l$. Primary entities received the highest weight, followed by secondary and additional entities. ORACLE aggregated these weighted scores at the case and patient levels and also reported category-level, level-specific and category-by-level micro and macro summaries. The principal ORACLE score was the weighted rubric-guided entity-coverage score.

\subsection{Training Recipe}

The training of OrthoPilot follows a structured curriculum that progressively transforms a 32-billion-parameter foundation model into a clinically agentic reasoning engine. The first three stages are conducted in a single-turn, tool-free setting, whereas the last two stages are conducted in a multi-turn, tool-augmented setting. This design allows the model to first acquire broad medical knowledge and musculoskeletal care reasoning ability in closed-context supervision and then learn evidence acquisition, tool use, and long-horizon decision consistency over full-pathway trajectories.

\subsubsection{Stage I: Clinical domain adaptation}

In Stage I, we adapted the base model to the medical domain through large-scale continued pre-training on high-quality medical corpora. This stage strengthened representations of medical terminology, disease mechanisms, diagnostic concepts and treatment semantics, providing a broad knowledge substrate for downstream musculoskeletal care reasoning.

\paragraph{Training data.}
The Stage-I corpus contained 6{,}565{,}111 medical examples and exceeded 10 billion tokens. It comprised publicly available medical text, question-answer pairs, clinical guidelines, biomedical literature, medical dialogue and interdisciplinary resources. Major sources included AGCT, Apollo, Clinical\_Guidelines and ShenNong-TCM-Dataset, with complete source details and sample counts provided in Supplementary Table~\ref{tab:dataset_availability}.

\subsubsection{Stage II: Musculoskeletal reasoning cold-start alignment}

After domain-level adaptation, Stage II aligned the model with musculoskeletal care reasoning through supervised fine-tuning on task-specific instruction data. This stage provided a reasoning-oriented cold-start policy, training the model to generate clinically grounded responses from patient evidence rather than short-form labels alone.

\paragraph{Training data.}
The core data source for this stage comprised 1{,}133{,}968 task-specific examples derived from the 21-year longitudinal cohort at Ruijin Hospital. These examples covered Tasks~1--11, with each question paired with the patient-specific context required for the corresponding decision point, including available imaging findings, laboratory results, operative information and longitudinal documentation. To preserve broader medical reasoning ability, we further mixed in external high-quality medical reasoning corpora. Teacher-generated rationales were retained only when the associated final answer remained clinically valid after automated filtering. The final Stage-II mixture contained 35{,}242{,}453 instruction instances.

\subsubsection{Stage III: Reason-driven reinforcement learning with GRPO}

Stage III further optimized reasoning quality through reinforcement learning, with the goal of improving response correctness, structural compliance and self-calibration under clinically uncertain settings. At this stage, training remained single-turn and tool-free. The model was rewarded solely on the quality of its generated reasoning and answer, without access to external tools or iterative retrieval.

We adopted Group Relative Policy Optimization (GRPO) \methodcite{shao2024deepseekmath}{52}, which compares multiple candidate responses sampled under the same prompt and computes relative advantages within the group. For each training prompt $x$, a group of responses $\{o_i\}_{i=1}^{G}$ was sampled from the old policy $\pi_{\theta_{\mathrm{old}}}$. The surrogate objective was:
\begin{equation}
\mathcal{J}_{\mathrm{GRPO}}(\theta)
=
\mathbb{E}_{\substack{x\sim P(\mathcal{X})\\
\{o_i\}_{i=1}^{G}\sim \pi_{\theta_{\mathrm{old}}}(\cdot\mid x)}}
\Biggl[
\frac{1}{G}
\sum_{i=1}^{G}
\min\!\left(
\begin{aligned}
&r_i(\theta)\hat{A}_i,\\[-0.15em]
&\operatorname{clip}\!\bigl(r_i(\theta),1-\epsilon,1+\epsilon\bigr)\hat{A}_i
\end{aligned}
\right)
\Biggr],
\end{equation}
where the policy ratio was:
\begin{equation}
r_i(\theta)=
\frac{\pi_{\theta}(o_i\mid x)}{\pi_{\theta_{\mathrm{old}}}(o_i\mid x)}
\end{equation}
and $\hat{A}_i$ denotes the group-relative advantage obtained by normalizing rewards within the sampled group.

\Needspace{8\baselineskip}
\paragraph{Reward design.}
The reward combined structural compliance and task accuracy:
\begin{equation}
R_i = R_i^{\mathrm{fmt}} + R_i^{\mathrm{acc}},
\end{equation}
where $R_i^{\mathrm{fmt}}$ penalized responses that violated the required output structure and $R_i^{\mathrm{acc}}$ evaluated answer correctness for verifiable QA targets. Exact matching was used when the target answer had a deterministic form, and LLM-as-a-judge assessment was used when correctness required semantic comparison with the reference answer.

\paragraph{Training data.}
We curated 49{,}275 high-value training examples for this stage, comprising musculoskeletal task examples and multidisciplinary scientific reasoning examples. The training set emphasized cases requiring deeper reasoning, ambiguity resolution and clinically consequential decision-making.

\subsubsection{Stage IV: Multi-horizon clinical trajectory distillation}

Stages I--III trained CHEESE in a tool-free regime. In contrast, Stage IV exposed the model to multi-turn, tool-augmented clinical reasoning for the first time. This stage trained the model to generate a task answer and to acquire the evidence needed to support that answer, first at the level of individual tasks and then across temporally ordered full-pathway decision chains.

\paragraph{Short-horizon trajectory supervision.}
Training began with short-horizon, task-specific trajectories, in which the model learned to use tools to solve one of the 11 musculoskeletal tasks under evidence constraints. A short-horizon trajectory was denoted by:
\begin{equation}
H^{\mathrm{short}} = (\tau_0, a_0, o_0, \ldots, \tau_T, a_T, y),
\end{equation}
where $\tau_t$ denotes the reasoning trace at step $t$, $a_t$ the selected action, $o_t$ the resulting observation, and $y$ the final task response.

The short-horizon supervision objective was written as a trajectory-level next-token prediction loss:
\begin{equation}
\begin{aligned}
\mathcal{L}_{\mathrm{short}}(\theta)
={}&-
\mathbb{E}_{H^{\mathrm{short}}\sim \mathcal{D}_{\mathrm{short}}}
\Biggl[
\sum_{t=0}^{T}\log P_{\theta}(\tau_t, a_t \mid H_{<t})
\\[-0.15em]
&\qquad+
\sum_{j=1}^{|y|}\log P_{\theta}(y_j \mid H_T, y_{<j})
\Biggr].
\end{aligned}
\end{equation}
Here, $\mathcal{D}_{\mathrm{short}}$ denotes the short-horizon tool-augmented supervised dataset. The first term supervises reasoning traces and tool calls, whereas the second term supervises the final answer. Tool-returned observations were treated as external context and did not contribute to the loss.

\paragraph{Long-horizon full-pathway trajectory supervision.}
After short-horizon distillation, training proceeded to long-horizon, full-pathway reasoning. The 11 tasks were concatenated into a temporally ordered clinical decision chain that mirrors inpatient care. At the start of a trajectory, the model received only the initial admission information. It then interacted with the in-hospital and external evidence worlds to acquire laboratory results, imaging findings, pathology, consultation records and guideline-level support as needed while solving the task sequence in chronological order.

The full-pathway trajectory for one case was denoted by:
\begin{equation}
\mathcal{H}^{\mathrm{path}} = \{H^{(1)}, H^{(2)}, \ldots, H^{(M)}\},
\end{equation}
where $M$ is the number of realized decision nodes in that patient case and each $H^{(m)}$ is the trajectory of the $m$th task. The long-horizon supervised loss was defined as a sum over task-level trajectory losses:
\begin{equation}
\mathcal{L}_{\mathrm{path}}(\theta) = \sum_{m=1}^{M} \mathcal{L}_{\mathrm{traj}}^{(m)}(\theta),
\end{equation}
where $\mathcal{L}_{\mathrm{traj}}^{(m)}(\theta)$ denotes the trajectory loss for the $m$th task node. This formulation trained the model to maintain temporal coherence across sequential clinical nodes. The joint distillation objective for Stage~IV combined both horizons:
\begin{equation}
\mathcal{L}_{\mathrm{IV}}(\theta) = \mathcal{L}_{\mathrm{short}}(\theta) + \lambda\, \mathcal{L}_{\mathrm{path}}(\theta),
\end{equation}
where $\lambda$ balances the relative weight of short-horizon and long-horizon supervision.

\paragraph{Training data synthesis.}
For short-horizon data, we started from high-quality musculoskeletal task instances and generated candidate multi-turn reasoning trajectories with tool-based evidence acquisition. Multiple frontier teacher models were used to sample diverse solution paths. Candidate trajectories were filtered by final-answer correctness and then ranked by an LLM judge that considered answer quality, tool-use appropriateness and problem-solving diversity. For long-horizon data, information that had previously been embedded in task-specific prompts was converted into explicit observations that the model had to retrieve. During synthesis of task $m+1$, the complete evidence-grounded trajectory from task $m$ was retained as part of the running context, so that downstream reasoning was conditioned on the accumulated patient history rather than on a reset prompt. Candidate full-pathway trajectories were sampled from multiple teacher models and filtered by strict rejection sampling. A pathway was retained only if all realized task answers along that trajectory were correct. The combined tool-augmented supervised dataset for Stage~IV contained approximately 157{,}000 trajectory records and 2.169 billion tokens.

\subsubsection{Stage V: Full-pathway agentic reinforcement learning}

In the final stage, CHEESE was optimized through full-pathway agentic reinforcement learning. Unlike Stage~III, which rewards single-turn reasoning outputs, Stage~V optimizes complete multi-turn trajectories in which the model must decide when to invoke tools, which evidence source to query and how to integrate returned observations before producing the final recommendation. This stage therefore reinforces both clinical answer quality and the action policy that governs evidence acquisition across the dual evidence worlds. Given a sampled trajectory $H_T$, the trajectory reward was defined as:
\begin{equation}
R(H_T)
=
\alpha R_{\mathrm{acc}}(H_T)
+
\beta R_{\mathrm{fmt}}(H_T)
+
\gamma R_{\mathrm{tool}}(H_T),
\end{equation}
where $R_{\mathrm{acc}}(H_T)$ is the task-accuracy reward, $R_{\mathrm{fmt}}(H_T)$ is the format-compliance reward, and $R_{\mathrm{tool}}(H_T)$ is the tool-use reward. The coefficients $\alpha$, $\beta$ and $\gamma$ balance answer correctness, structural compliance and appropriate tool invocation, respectively.

The corresponding agentic reinforcement-learning objective was:
\begin{equation}
\mathcal{J}_{\mathrm{Agentic}}(\theta)
=
\mathbb{E}_{(q,c)\sim \mathcal{D}_{\mathrm{RL}},\, H_T\sim p_{\theta}(\cdot\mid q,c,\mathcal{T})}
\left[
R(H_T)
\right],
\end{equation}
where $\mathcal{D}_{\mathrm{RL}}$ denotes the reinforcement-learning distribution over task queries and patient contexts, $\mathcal{T}$ denotes the available Tool Plaza interfaces, and $p_{\theta}(H_T\mid q,c,\mathcal{T})$ denotes the trajectory distribution induced by the CHEESE policy and tool-returned observations.

\paragraph{Training data mixture.}
The reinforcement-learning stage used both single-task queries from the 11 musculoskeletal tasks and full-pathway continuation queries. In continuation queries, the model received the evidence-grounded trajectories of earlier tasks and was trained to solve the next decision node. This mixed design strengthened both local decision accuracy and temporal coherence across the care pathway. We curated high-value training examples for this stage and performed distributed optimization under GRPO-style updates.

Overall, the CHEESE training curriculum represents a transition from next-token prediction to clinical execution. By progressively moving from medical knowledge adaptation to musculoskeletal care reasoning cold-start alignment to multi-horizon clinical trajectory distillation and full-pathway agentic reinforcement learning, the model acquired domain expertise, tool-use competence and evidence-grounded temporal coherence across the full musculoskeletal care pathway.

\newpage

\section*{Code Availability}
Our source code is available at \url{https://github.com/SII-WenjieLisjtu/OrthoPilot}.

\section*{Data Availability}
Figures were generated and processed using Python with Matplotlib and seaborn. Schematic illustrations were created using BioRender (\url{https://biorender.com}). Publicly available datasets used for model training are listed in Supplementary Table~\ref{tab:dataset_availability}. Public repository links include:
\begin{itemize}
    \item Apollo: \url{https://huggingface.co/datasets/FreedomIntelligence/ApolloCorpus}
    \item biorxiv-clustering-s2s: \url{https://huggingface.co/datasets/mteb/biorxiv-clustering-s2s}
    \item ChatMed-Consult-Dataset: \url{https://huggingface.co/datasets/michaelwzhu/ChatMed_Consult_Dataset}
    \item gliner-bird-diet-synthetic: \url{https://huggingface.co/datasets/wjbmattingly/gliner-bird-diet-synthetic}
    \item med\_qa: \url{https://huggingface.co/datasets/bigbio/med_qa}
    \item Medical\_Cord19: \url{https://huggingface.co/datasets/allenai/cord19}
    \item medqa\_corpus\_en: \url{https://huggingface.co/datasets/cogbuji/medqa_corpus_en}
    \item ReMeDi: \url{https://huggingface.co/datasets/ljvmiranda921/remedial-xsd-sft}
    \item ShenNong-TCM-Dataset: \url{https://huggingface.co/datasets/michaelwzhu/ShenNong_TCM_Dataset}
\end{itemize}
Other public or synthetically generated resources are described in Supplementary Table~\ref{tab:dataset_availability}. The Ruijin Hospital clinical records and institutionally derived datasets used in this study are subject to participant confidentiality, institutional ethics approval and data-protection regulations. To protect patient privacy, these data are not openly released. Access for scientific research can be requested through a controlled access process by submitting a detailed research proposal and Institutional Review Board approval from the applicant's home institution to the Ruijin Hospital data access committee. Approved data use will be limited to the scope of the reviewed research proposal and will require appropriate data-use agreements. Data generated during this study, including the main results and training and testing procedures, are available within this Article and its Supplementary Information. Source data supporting the aggregate analyses are provided with this paper.

\renewcommand{\refname}{Methods References}
\renewcommand{\bibname}{Methods References}
\begingroup
\makeatletter

\endgroup

\section*{Acknowledgements}

This work was supported by the National Key Research and Development Program of China (2023YFC2410705 and 2023YFC2410700), the National Natural Science Foundation of China (82472419 and 82502877), the Shanghai Municipal Science and Technology Commission Project (25ZR1402348) and Shanghai Artificial Intelligence Laboratory. We thank all participating patients and medical and technical staff for their valuable contributions; G. Ning for institutional support and constructive discussions; Shanghai Innovation Institute for providing computational resources; J. Li for rigorous oversight of ethical principles; M. Zhang for valuable comments during manuscript revision; and Q. Lu of MiraclePlus and X. Wang of Baichuan Intelligent Technology for valuable scientific advice and technical support.

\section*{Author Contributions}

W.L., X.W. and Lei Wang conceived the idea. W.L. designed the overall research framework, defined the longitudinal clinical-management tasks, designed the study and evaluation framework, organized data curation and case review, coordinated system development and experimental evaluation, performed the main analyses, prepared the figures and wrote the paper. W.L., Yujie Zhang, F.Z. and H.S. led the core computational algorithm design, model training, system development, experimental evaluation and technical extensions, and contributed to figure preparation and manuscript writing. R.Y. contributed to clinical data curation, case review and medical content verification. J.H., W.H., Yuanfeng Ji and X.H. contributed to methodological discussions, task modelling, model-evaluation design and optimization of the technical strategy. Y.C., Lilong Wang and Yankai Jiang implemented data processing, data structuring and quality control of model inputs and outputs. C.W., L.C., K.Z., C.M. and Jiawei Liu contributed to model implementation, system-module development and construction of the experimental workflow. Jiyao Liu, M.H. and H.G. performed experimental execution, system testing and technical verification. K.W., L.Y. and Y.D. contributed to technical evaluation, analysis of experimental results and interpretation of computational findings. H.W., A.L.W., Y.H. and S.Y. contributed to clinical data preparation, case screening and organization of clinical materials. G.L., Y.X., C.Z. and Yin Zhang contributed to case review, expert annotation, clinical-consistency assessment and medical interpretation. L.L. and C.G.S. provided clinical interpretation of outcomes, medical validation, assessment of clinical feasibility and advice on clinical translation. Y.Q., W.M. and X.W. provided guidance on algorithm design, AI-agent architecture, experimental strategy, technical infrastructure and computational resources. Lei Wang provided guidance on clinical study design, the clinical-management framework, medical interpretation and clinical translation. Lei Wang, Y.Q., W.M. and X.W. supervised the research. All authors discussed the ideas and results, contributed to the drafting and revising of the manuscript, approved the final version for submission, and agree to be accountable for all aspects of the work, ensuring that questions related to the accuracy or integrity of any part of the work are appropriately investigated and resolved.

\section*{Competing interests}
The authors declare no competing interests.

\end{document}